\RequirePackage{fix-cm}
\documentclass[smallcondensed]{svjour3}     
\smartqed  
\usepackage{graphicx}
\usepackage{tikz}
\usetikzlibrary{bayesnet}
\usepackage{amsmath}
\usepackage{amsfonts}
\usepackage{amssymb}
\usepackage{tikz}
\usetikzlibrary{arrows,matrix,positioning,backgrounds,trees}
\usepackage{algorithm}
\usepackage{algorithmic}
\usepackage{tablefootnote}

\usepackage{graphicx}

\usepackage{verbatim}

\usepackage{multirow}
\usepackage{longtable}

\usepackage{booktabs}
\usepackage{url}

\newcommand{\oneh}[1]{H${}^{#1}$\thinspace}
\newcommand{\fifteenn}{N\thinspace}
\newcommand{\thirteenc}[1]{C${}^{#1}$\thinspace}
\newcommand{\aH}{\text{H}}
\newcommand{\aN}{\text{N}}
\newcommand{\aC}{\text{C}}
\newcommand{\thirteenco}{CO\textsc{\char13} }

\newcommand{\mbf}[1]{\mathbf{#1}}
\newcommand{\mcal}[1]{\mathcal{#1}}

\newtheorem{defn}{Definition}

\newtheorem{prob}{Problem}
\begin{document}

\title{NMR Assignment through Linear Programming
}


\author{Jos\'{e} F. S. Bravo-Ferreira         \and
        David Cowburn \and
        Yuehaw Khoo \and
        Amit Singer
}


\institute{Jos\'{e} F. S. Bravo-Ferreira \at
              PACM, Princeton University, NJ 08540 \\
              \email{josesf@princeton.edu}
           \and
           David Cowburn \at
             Departments of Biochemistry and of Physiology and Biophysics, Albert Einstein College of Medicine, NY 10461 \\
             \email{cowburn@cowburnlab.org}
           \and
           Yuehaw Khoo \at
              Department of Statistics, University of Chicago, IL 60637 \\
              \email{ykhoo@uchicago.edu}
           \and
           Amit Singer \at
           	  Department of Mathematics and PACM, Princeton University, NJ 08540 \\
              \email{amits@math.princeton.edu}
}

\date{Received: date / Accepted: date}

\maketitle

\begin{abstract}
Nuclear Magnetic Resonance (NMR) Spectroscopy is the second most used technique (after X-ray crystallography) for structural determination of proteins. A computational challenge in this technique involves solving a discrete optimization problem that assigns the resonance frequency to each atom in the protein. This paper introduces LIAN (LInear programming Assignment for NMR), a novel linear programming formulation of the problem which yields state-of-the-art results in simulated and experimental datasets.
\end{abstract}

\section{Introduction}\label{intro}

We investigate a type of constraint satisfaction problem on certain graphs arising in NMR Spectroscopy. Crucial to NMR spectroscopy is the time-consuming chemical shift assignment problem (also known as the spectral or resonance assignment problem) \cite{donald-book}, which inhibits the wider application of this technique. To date, this procedure is done largely in a semi-manual way, even though approaches using exhaustive search \cite{Mars}, integer programming \cite{ass:ipass}, genetic algorithms \cite{ass:flya}, variants of belief propagation \cite{ass:pine}, among others, have all shown promise in different experimental datasets. However, these approaches typically lack either a principled definition of the cost function, or a way to determine whether the global optimizer is every attained. In this paper, we attempt to address these issues in the search for a more rigorous algorithm.

\subsection{The assignment problem}

The spectral assignment problem is the problem of determining the resonance frequencies of individual atoms in the protein. These frequencies are typically defined by their chemical shifts, measured in parts per million (ppm) relative to a reference compound since they depend on the local environment of individual nuclei \cite{CAVANAGH}. Therefore, the resonance frequencies are often referred to as \emph{chemical shifts}. 

\begin{figure}[!h]
	\centering
	\includegraphics[width=\columnwidth]{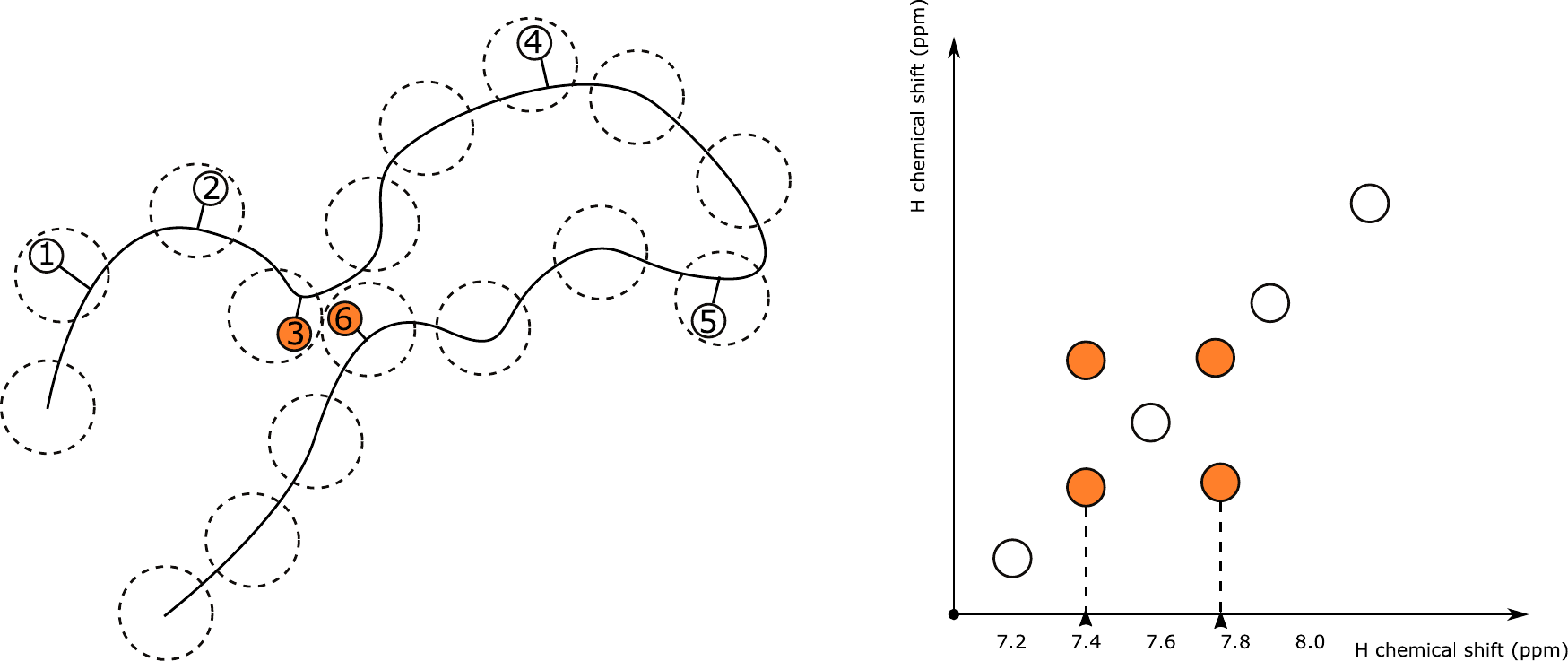}
	\caption[Illustration of \oneh{}-\oneh{} NOESY]{Illustration of a protein with two hydrogen atoms in close spatial proximity (left), which induce off-diagonal peaks in \oneh{}-\oneh{} NOESY spectrum (right).}
	\label{fig:1h-1h-noesy}
\end{figure}

To extract constraints from which one can deduce a global protein structure, NMR spectroscopists make use of interactions between atom nuclei, such as the nuclear Overhauser effect (NOE), among others. This effect arises from the dipolar relaxation of a two-spin system, and manifests itself as an off-diagonal peak in NOE spectroscopy experiments (NOESY) \cite{CAVANAGH}, as illustrated in Figure~\ref{fig:1h-1h-noesy}.

The NOE between two hydrogens (H) depends on the distance \cite{CAVANAGH}, such that cross-peaks in H-H NOESY spectra are indicative of the existence of two hydrogen atoms within close proximity. However, this information is not immediately useful geometrically without the knowledge of which hydrogen atoms induce the cross-peak. The assignment problem provides this information, by mapping the chemical shifts observed in this and other NMR spectroscopy experiments to the corresponding atoms in the protein. The assignment of all hydrogen and other \emph{backbone} atoms, is a crucial first-step for high-resolution structure determination in NMR \cite{lu-yu-lian}. 

\begin{figure}[]
	\centering
	\includegraphics[width=0.8\columnwidth]{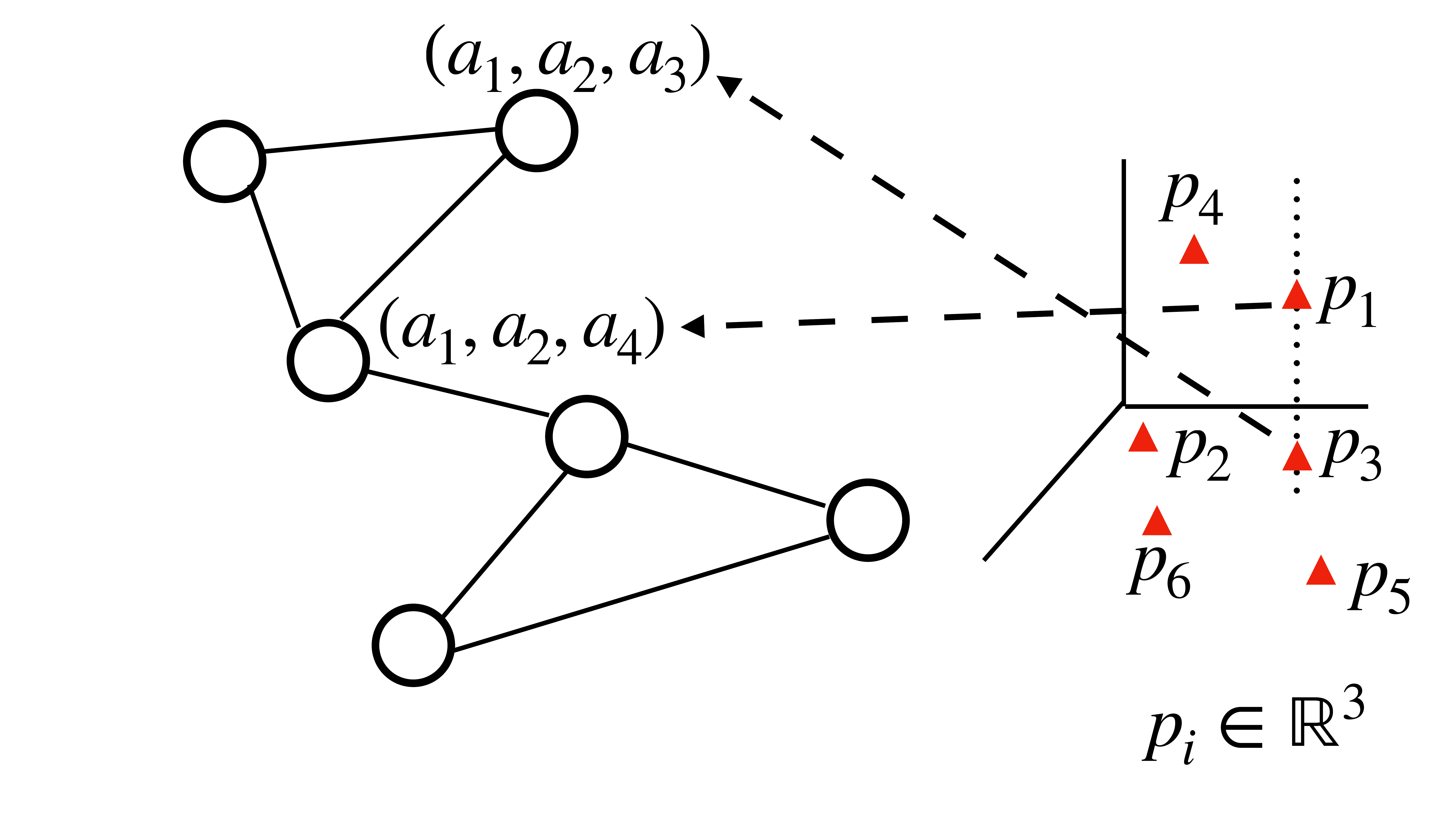}
	\caption{Representing the assignment problem as a graph. Each node is associated with a triplet of atoms (e.g. $(a_1,a_2,a_3)$ or $(a_1,a_2,a_4)$). When two triplets share some atoms, there is an edge joining them. Each triplet gives rise to a resonance peak in the 3-dimensional spectra. The goal is to assign the measured peaks to the nodes or triplets. Crucially, when two peaks result from two triplets that share some common atoms, they will share certain coordinates. For example, $(a_1,a_2,a_3)$ and $(a_1,a_2,a_4)$ share $(a_1,a_2)$, so the coordinates in the horizontal plane of the resulting peaks $p_1, p_3$ are the same (indicated by the vertical dotted line).}
	\label{fig:graph_rep_assignment}
\end{figure}

The process of NMR assignment (especially for larger proteins) relies on a set of experiments known as heteronuclear resonance experiments. Before describing these experiments in detail, we first provide some background information on protein structure. A protein is composed of a chain of \emph{residues}. Every residue contains the same set of atoms \oneh{N}, N, $\text{C}^\alpha$, $\text{C}^\beta$ (with the exception of Proline). These repeated elements then form the \emph{protein backbone}. Basic heteronuclear experiments couple (\oneh{N}, N), (\oneh{N}, N, $\text{C}^\alpha$), or (\oneh{N}, N, $\text{C}^\beta$) from a single residue or two adjacent residues. Ideally, these pairs or triplets contribute to the resonance peaks on a 2- or n-dimensional spectra, where the coordinates of the peak are the resonance frequencies of the hydrogen, nitrogen and carbon (similar to the case in Figure~\ref{fig:1h-1h-noesy}). As different triplets (or pairs) may share common atoms, this results in a graph such as the one depicted in Figure~\ref{fig:graph_rep_assignment}, where a node resembles a triplet (pair), and an edge between two nodes means two triplets (pairs) share one or two atoms. \emph{The goal of the assignment procedure is to take the measured peaks in $\mathbb{R}^3$ (peaks in $\mathbb{R}^2$ resulted from (N, \oneh{N}) can be embedded into $\mathbb{R}^3$), and assign them to the appropriate nodes in the graph}. An edge between two nodes induces a constraint that the two assigned peaks must share coordinates across certain dimensions.

In the next subsection, we describe different kinds of measurements one can perform to couple different pairs or triplets, giving rise to peaks in 2- or 3-dimensional spectra that facilitate the assignment procedure. Readers unfamiliar with the chemistry involved may skip the rest of the next subsection, and read Section~\ref{section:assignment} where we elucidate the general philosophy of how these spectra can be used in graph theoretic notions.

\subsection{Typical spectra used for assignment}\label{section:spectra}
In this section we detail three basic experiments which are commonly used for backbone assignment of small proteins ($<$150 residues). As we shall see, these three sets of experiments can provide an assignment of the peaks. They give rise to the three types of spectra detailed below.

\begin{figure}[]
	\centering
	\includegraphics[width=0.8\columnwidth]{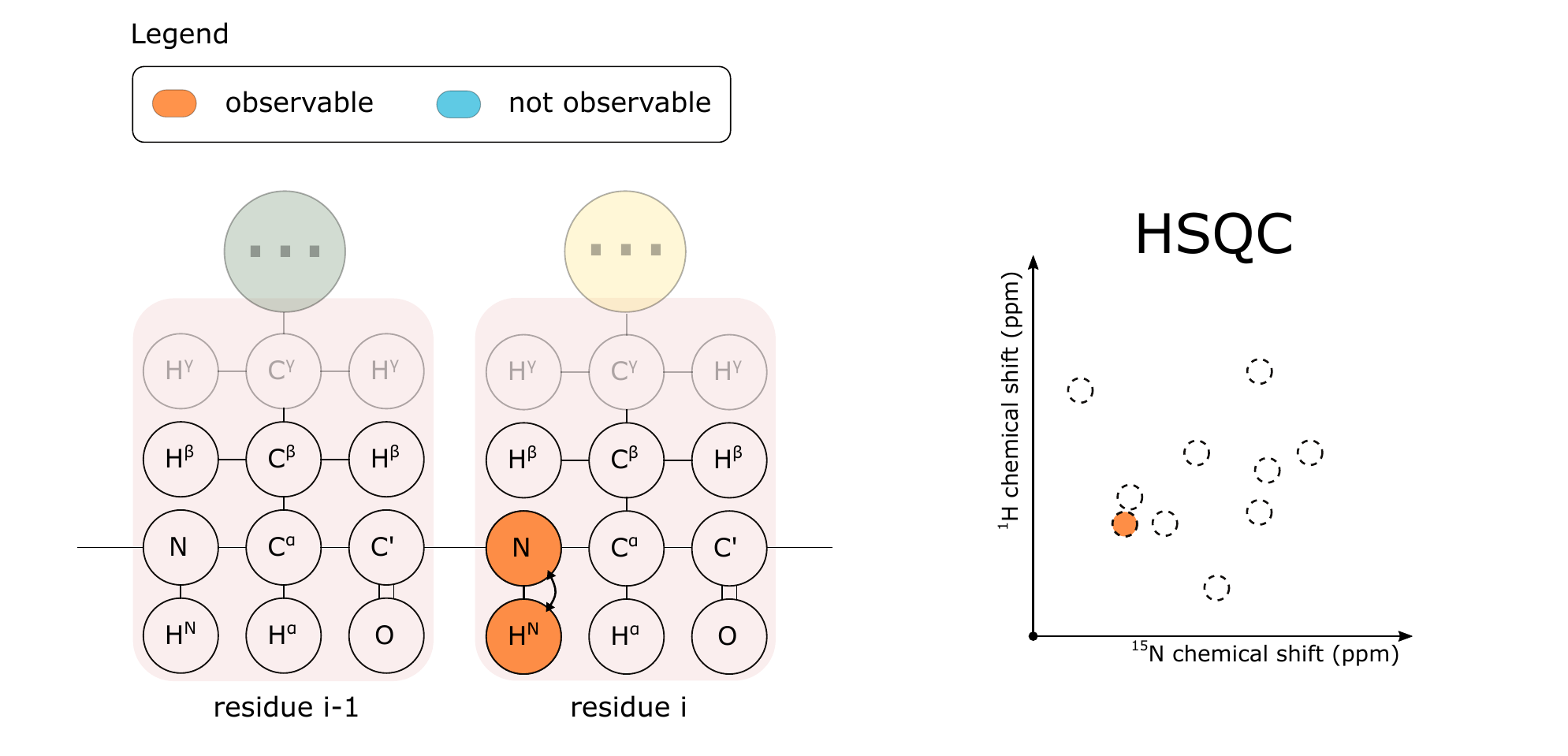}
	\includegraphics[width=0.8\columnwidth]{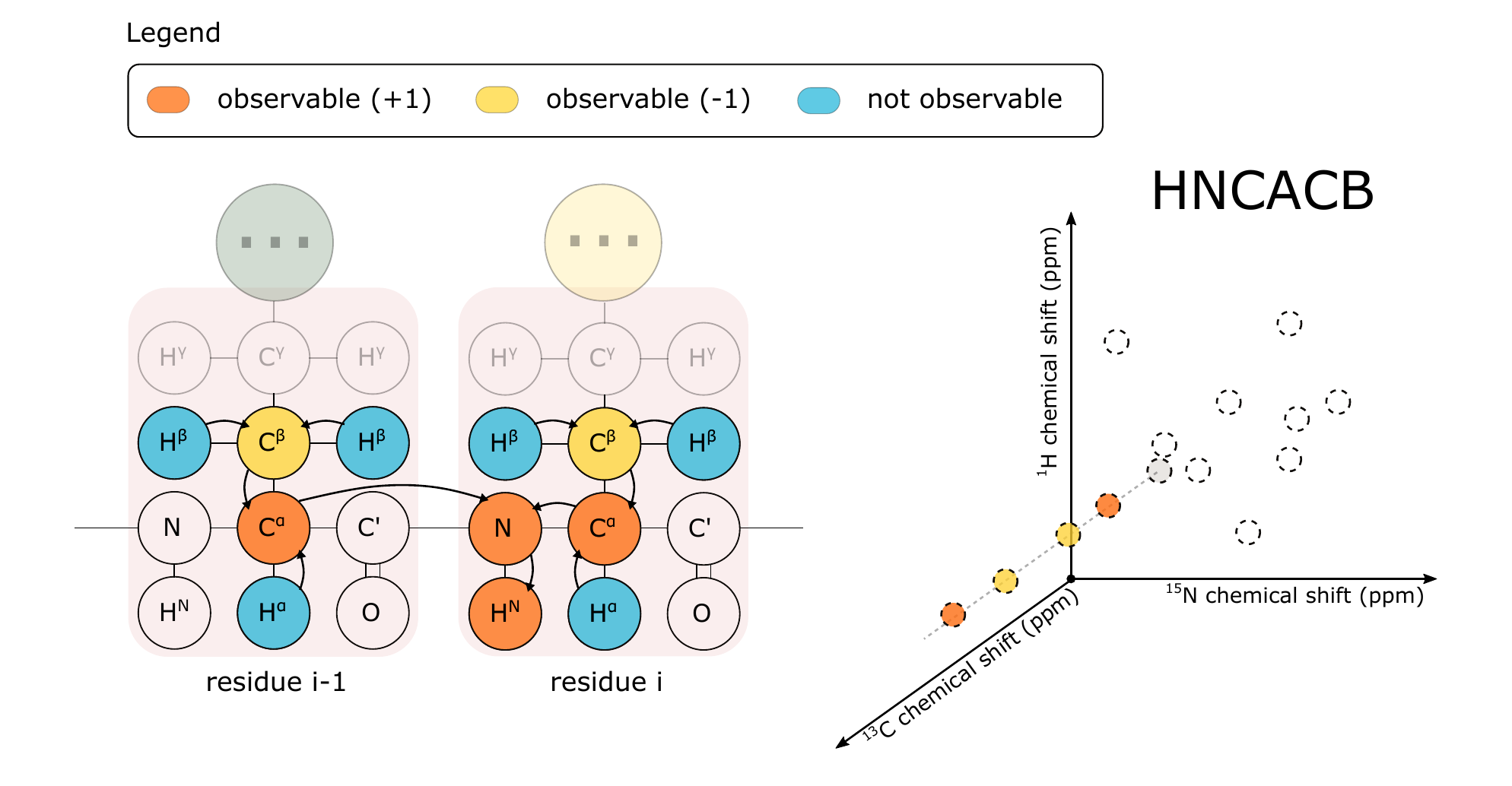}
	\includegraphics[width=0.8\columnwidth]{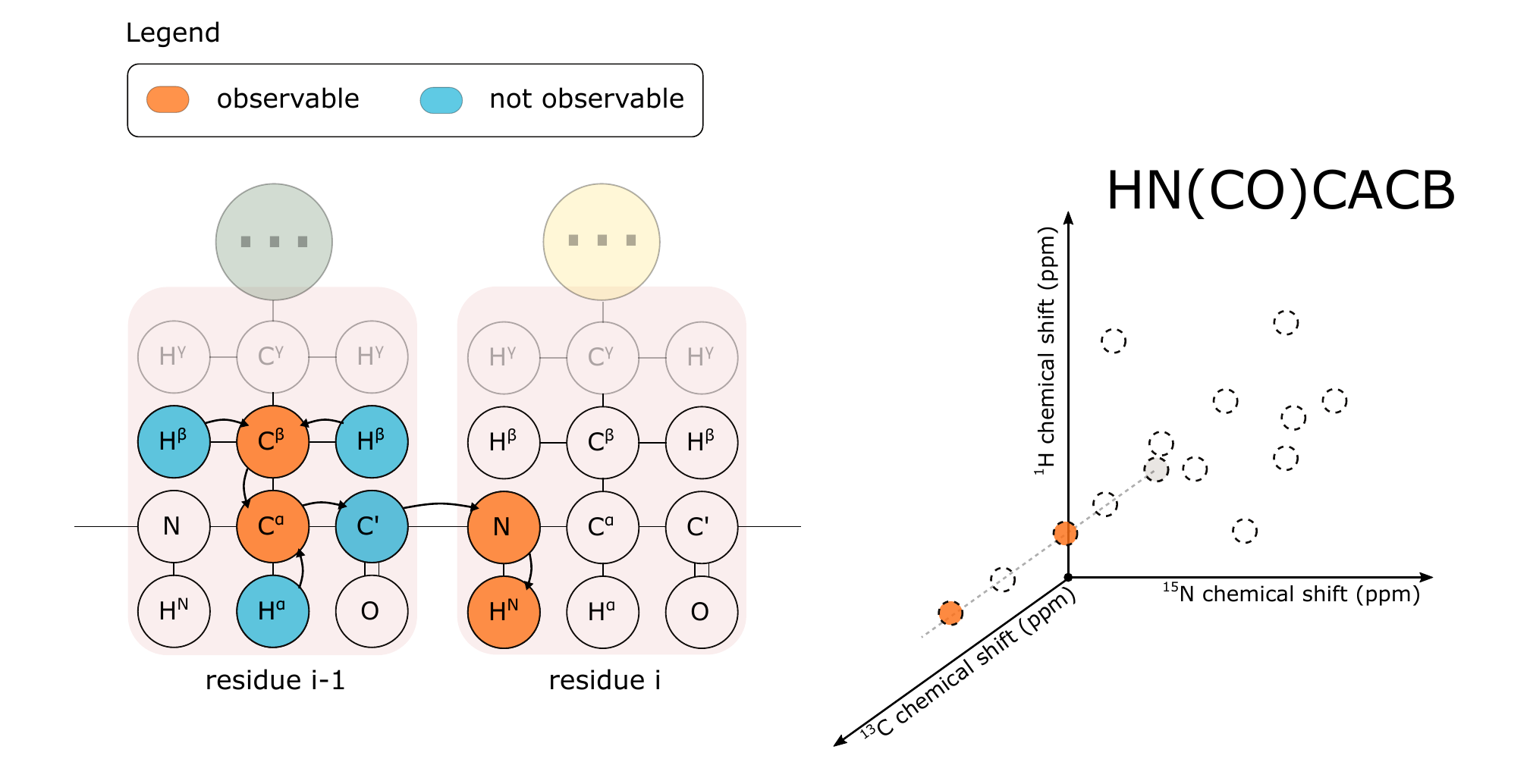}
	\caption[HSQC, HNCACB, HN(CO)CACB spectra for backbone assignment.]{Illustration of three heteronuclear spectra for backbone assignment. HSQC is used as a fingerprinting experiment. Peaks in HNCACB and HN(CO)CACB develop off the \fifteenn--\oneh{N} plane, along the carbon dimension. Polarity differences of antiphase peaks in HNCACB disambiguate \thirteenc{\alpha} from \thirteenc{\beta}, and HN(CO)CACB further disambiguates intra- from inter-residue atoms.}
	\label{fig:3-spectra}
\end{figure}

\paragraph{HSQC} The heteronuclear single quantum coherence experiment \cite{intro-hsqc}  involves a transfer of magnetization between the base amide proton, \oneh{N}, and the nitrogen \fifteenn and back, as illustrated in Figure \ref{fig:3-spectra}. With the exception of proline, all basic amino acids feature this amide pair, such that a distinct peak can be expected for most residues, leading to the use of HSQC as a fingerprinting experiment.

\paragraph{HNCACB} This experiment involves magnetization transfer from \oneh{\alpha} and \oneh{\beta} to \thirteenc{\alpha} and \thirteenc{\beta}, respectively, and then from \thirteenc{\beta} to \thirteenc{\alpha} and finally to \fifteenn and to \oneh{N} of the same or subsequent residue, as illustrated in Figure \ref{fig:3-spectra}, and described in \cite{intro-hncacb}. The polarities of the \thirteenc{\alpha} and \thirteenc{\beta} peaks are opposite, which allows these to be distinguished. Importantly, note that \thirteenc{\alpha} and \thirteenc{\beta} peaks are observed with the same root \fifteenn--\oneh{N} pair. This means there should be four peaks in HNCACB spectra, having the same frequency in the \fifteenn and \oneh{N} dimension. This allows for \emph{sequential walking} (Section \ref{section:seq walking}), the process of matching residues with their neighbors through matching carbon frequencies.

\paragraph{HN(CO)CACB} The last of the experimental toolset for backbone assignment of medium-size proteins also gives rise to \thirteenc{\alpha} and \thirteenc{\beta} peaks \cite{intro-hn(co)cacb}, as illustrated in Figure \ref{fig:3-spectra}. Magnetization transfer happens from \oneh{\alpha} and \oneh{\beta} to \thirteenc{\alpha} and \thirteenc{\beta}, onto \thirteenco and finally the base amide pair. Chemical shifts are evolved only on \thirteenc{\alpha} and \thirteenc{\beta} before detection, so no \thirteenco peaks are observed.

\subsubsection{Basic assignment procedure}\label{section:seq walking}
\begin{figure}[h]
	\centering
	\includegraphics[width=\columnwidth]{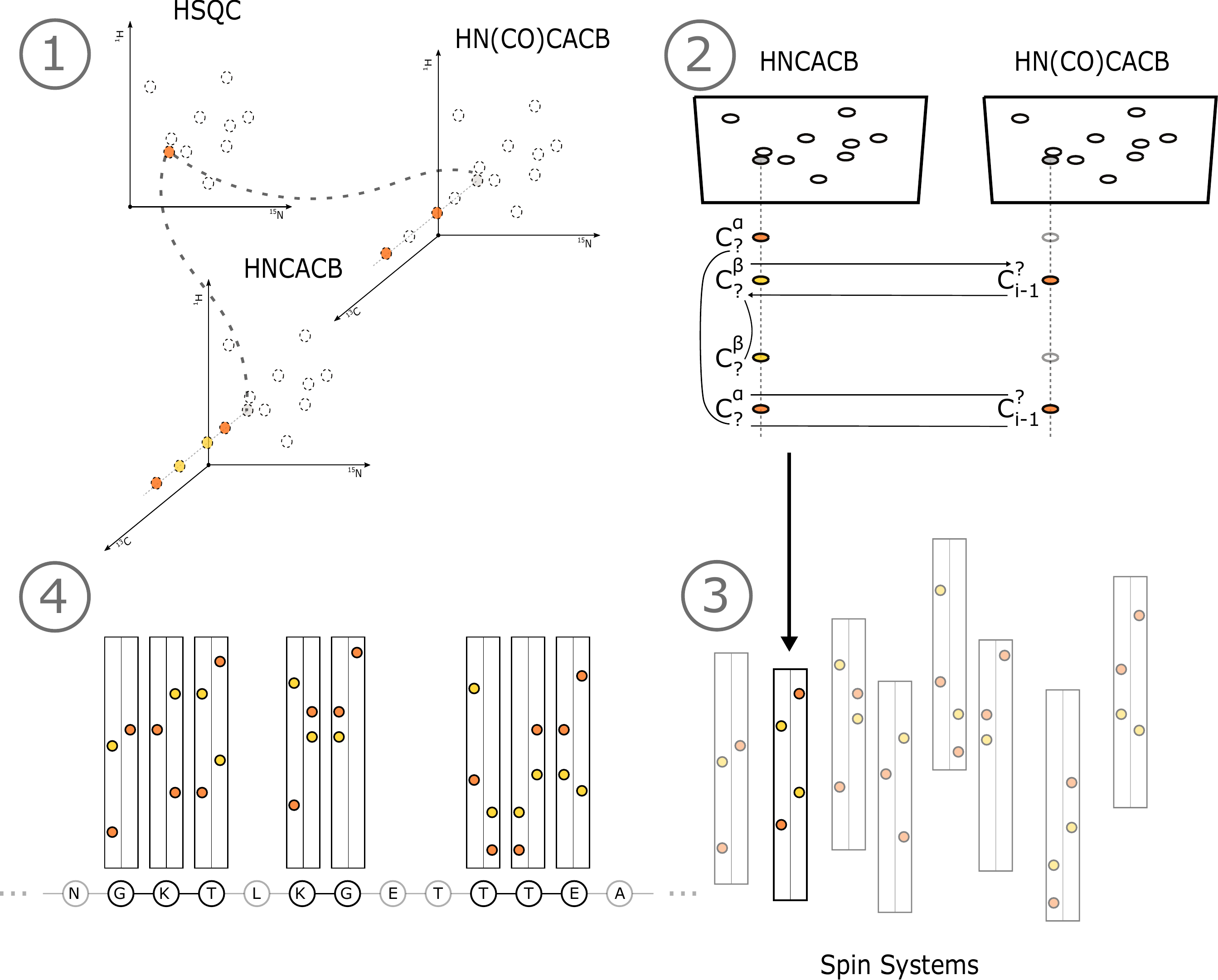}
	\caption[Sketch of backbone assignment procedure through heteronuclear NMR.]{Sketch of backbone assignment procedure through heteronuclear NMR. (1) HSQC peaks are used as fingerprints and linked to matching peaks in HN(CO)CACB and HNCACB spectra. (2) The carbon dimension in HNCACB and HN(CO)CACB is used along with phase information to identify specific atom frequencies. (3) Spin systems are created from these fragments, each containing carbon frequencies for two adjacent residues. (4) Spin systems are ordered into fragments (based on matching carbon frequencies) and placed in their correct position through statistical typing (using prior information from a public database of chemical shift statistics).}
	\label{fig:assignment}
\end{figure}

Given this set of experiments, a greedy way of assignment (Figure  \ref{fig:assignment}) is summarized in this section. This procedure forms the backbone of many assignment algorithms. It is as follows:

\begin{enumerate}
	\item[(1)] HSQC is used as a fingerprint experiment due to high sensitivity and resolution, allowing for accurate determination of base \fifteenn--\oneh{N} pairs. As we can see in Figure \ref{fig:assignment}, peaks in HNCACB and HN(CO)CACB can be grouped according to frequency in the \fifteenn and \oneh{N}. Therefore, peaks in HSQC are matched with peaks in HNCACB and HN(CO)CACB spectra which satisfy tolerance bounds (typically 0.02 - 0.03 ppm for hydrogen and 0.20 - 0.30 ppm for nitrogen). 
	
	\item[(2)] After grouping the peaks in HNCACB and HN(CO)CACB, within the same \fifteenn--\oneh{N} grouping, the peaks are further correlated and disambiguated using phase information, allowing for the assembly of spin systems. Firstly, HN(CO)CACB tells which of the four peaks in HNCACB comes from the carbons of previous residue, while the +/- sign in HNCACB distinguishes \thirteenc{\alpha} from \thirteenc{\beta}.
	
	\item[(3)] After steps (1) and (2), peaks from HSQC, HN(CO)CACB, HNCACB are combined, resulting in groups of peaks where each group has four peaks. Each group is called a \emph{spin system}. We re-emphasize that the peaks within the same spin system have the same \fifteenn and \oneh{N} frequency, but the frequency along the carbon axis differs. There should be as many spin systems as the number of residues (with a few exceptions), since each residue has exactly one pair of \fifteenn--\oneh{N}.
	
	\item[(4)] Now we want to associate the spin systems to the residues in the protein. As depicted in Figure \ref{fig:assignment}, if two spin systems come from two adjacent residues, they share two peaks with the same carbon chemical shifts. This gives a criteria to create fragments through sequential walking along the \thirteenc{\alpha} and \thirteenc{\beta} chains. Now the fragments should come from a certain segment of residues in the protein chain. This is done via \emph{statistical typing}, which compares the measured chemical shifts of atoms in the identified fragments with the expected chemical shift of the residues collected in a public database such as BMRB \cite{ass:BMRB}. The fragments are placed optimally according to that prior.
\end{enumerate}

We remark that that the widespread availability of NMR data collected in databases such as BMRB is fundamental in assignment. The distributions of chemical shifts in different amino acid types is not the same, due to the unique environment induced by the different chemical structures. Certain amino acids, including alanine, glycine, isoleucine, leucine, proline, serine, threonine, and valine, present particularly distinct signatures. Among these, Glycine and Proline are unique in that they do not feature certain peaks. Specifically, Glycine is characterized by its unique \thirteenc{\alpha} shift and the absence of a \thirteenc{\beta} signal (see, e.g. \cite{Glycine93}). Proline, as the only secondary amine among proteinogenic amino acids, does not yield peaks in the experiments described above. As a result, such residues are only identified as neighbors through HN(CO)CACB spectra or via specific experiments (e.g. \cite{Proline20}).

We also note, however, that the local chemical environment can shift the resonance frequencies of certain atoms, even if they belong to the same residue type. In fact, local chemical shifts can be used as predictors for the chemical shift of a specific atom (see, e.g., \cite{HASH}), which means that sophisticated probabilistic descriptions of the resonance frequencies may be necessary for certain proteins, or that an iterative process taking into account the primary and secondary structure of the protein should be employed.

\subsubsection{Challenges in sequential assignment}

As described above, accurate assignment relies on the correct identification of peaks in NMR experiments, and their assembly into consistent spin systems that can be sequentially assigned.

In practice, as the quality of NMR spectra deteriorates, some peaks will overlap, and others cannot be detected at all, due to peak broadening and lower SNR. Artifacts included in automatically selected peak lists further hamper sequential assignment. Even with a decent set of spin systems, sequential assignment itself is not as simple as solving a one-dimensional puzzle, as experimental noise, erroneous spin systems, overlapping chemical shifts, and missing spin systems introduce ambiguity to the process.

\subsection{Our contributions}
The contribution of the paper is two-fold:
\begin{enumerate}
\item We formulate the spectral assignment problem as a constraint satisfaction problem, with cost being defined on a graph $G=(V,E)$:
\begin{equation}
\label{eq:general graphical MLE}
\min_{\{z_i\}_{i\in V} \subset \{0,1\}^m}\sum_{(i,j)\in E} f_{ij}(z_i,z_j),\quad \text{s.t.} \ \text{linear constraints on}\ z_1,\ldots,z_{\vert V \vert}.
\end{equation}
Here $z_i$'s are indicator vectors associated with nodes $V$.
\item In general, this type of problem is hard to solve, unless $G$ is tree-like. However, the adjacency matrix of graph $G$ in the assignment problem forms a band matrix, which allows us to reformulate \eqref{eq:general graphical MLE} as a problem on a path graph, by clustering the nodes in $G$. Such a reformulation allows \eqref{eq:general graphical MLE} to be solved either via dynamic programming or linear programming (LP), depending on the structure of the constraints.
\end{enumerate}

\subsection{Organization}
In Section \ref{section:assignment}, we formulate the assignment problem as a constraint satisfaction problem over discrete domain. In Section \ref{methodology}, we present a few versions of LPs to solve the constraint satisfaction problem. In Section \ref{results}, we demonstrate the performance of our algorithm in simulated and experimental datasets. However, we begin by surveying a few works that are most relevant to the proposed method.

\subsection{Prior work}\label{intro:existing-work}

As early as 2004, a detailed review identified twelve important works on automated NMR assignment \cite{ass:review2004}. A more recent protocol overview \cite{ass:review2012} cited 44 works on automated chemical shift assignment, which is still not a complete list. Nearly all of the works cited leveraged a similar pipeline of: (1) registering peaks across different dimensions, (2) spin system construction, (3) fragment building through sequential walking, and finally (4) mapping of fragments through probabilistic typing, where a variety of different techniques have been explored, including exhaustive search \cite{ass:mapper}, best-fit heuristics \cite{ass:autoassign}, simulated annealing/monte carlo \cite{ass:donald-sa}, \cite{ass:monte}, \cite{ass:pasta}, and genetic algorithms \cite{ass:garant}, \cite{ass:nsga}, \cite{ass:match}. Among these, only a small subset has seen extensive use reported on the protein data bank (PDB, \cite{ass:PDB}) including AutoAssign \cite{ass:autoassign}, CYANA  \cite{ass:cyana}, GARANT \cite{ass:garant}, and PINE \cite{ass:pine}. This highlights how automated assignment techniques have not yet managed to achieve widespread adoption. 

The development of new automated assignment tools is challenging on a technical level, but there are additional barriers that must be mitigated. Namely, it is currently challenging to fairly compare assignment algorithms, due to discrepancies in input formats, simulation assumptions, and the lack of reproducibility standards or benchmarking datasets. Many state-of-the-art tools, such as CYANA \cite{ass:cyana} (or FLYA, \cite{ass:flya}, which is available as part of the CYANA package) lie behind a pay wall. Benchmark datasets are rarely open sourced, such that reliable comparisons can only be made through simulations. Since simulation code is rarely open sourced, comparisons require replicating the simulation frameworks adopted in other works, which is time consuming and error prone.

In \cite{Cisa}, the authors attempted to rectify this by introducing a standardized simulated test suite of spin systems, produced according to empirically accepted experimental noise margins. The authors tested their algorithm against three other assignment tools: an iterative, connectivity-based approach called PACES \cite{ass:paces}, the random-graph theoretic approach RANDOM \cite{ass:RANDOM}, and MARS \cite{Mars}, yet another iterative, connectivity-based method using random permutations to progressively nudge assignments into better ones. An integer programming approach called IPASS \cite{ass:ipass} later tested on this same experimental suite. We include comparisons on this same test suite for our algorithm.

Out of all automated assignment methodologies, our approach shares the most similarities with IPASS and FLYA. These are all algorithms that adopt a global optimization view of the assignment problem, rather than optimizing locally through fragment building. We describe them briefly below.

\paragraph{IPASS:} This algorithm begins with a graph-based procedure to build spin systems from peak lists of HSQC, HNCACB, and HN(CO)CACB spectra \cite{ass:ipass}. Distances between peaks are calculated based on the chemical shifts of carbon, nitrogen, and hydrogen atoms, with any distances smaller than twice the nearest-neighbor distance converting into an edge. Spin systems are obtained through brute-force search of consistent $\aC^\alpha$ and $\aC^\beta$ values. A connectivity graph is established by creating edges between any two spin systems where the $\aC^\alpha$ and $\aC^\beta$ connections satisfy a loose threshold of $\delta = 0.5\text{ppm}$, and a heuristic connectivity score is computed for each edge, with edges scoring below a threshold score trimmed from the graph. Finally, all combinations of fragments where no spin system appears in more than one fragment and no fragments overlap are enumerated. An integer linear program is then solved for each such combination to compute an assignment that best agrees with a probabilistic prior on the frequencies of each protein residue.

\paragraph{FLYA:} Unlike IPASS, FLYA attempts to optimize a global score directly from peak lists, without the intermediate steps of spin system construction or fragment building \cite{ass:flya}, with state-of-the-art results. Given a set of measured peak lists, FLYA compares it directly to a hypothetical set of peak lists which one would expect to observe given the NMR experiments that were carried out. Expected peaks are matched to measured peaks with the goal of maximizing the global score, and chemical shift values are inferred from this matching. The score is based on a likelihood computed from a generative model assumed to produce the experimental data (we refer the reader to \cite{ass:flya} for details). The optimization process itself is a combination of heuristic local optimizations (which remap local regions of the assignment) and a genetic algorithm, which probabilistically recombines and mixes existing assignments from generation to generation.

\subsubsection{A note on generative assumptions}

While probabilistic assumptions are commonly made across automated assignment tools, they do not always coincide. In particular, CISA \cite{Cisa} and, by extension, IPASS \cite{ass:ipass} generate simulated data by adding white noise chemical shifts at the spin system level. FLYA, on the other hand, assumes truncated Gaussian noise on measured peaks to ensure valid assignments are possible under their evaluation framework \cite{ass:flya}. As a result, great care is required when comparing different algorithms.

\section{Spectral assignment problem as constraint satisfaction problem}\label{section:assignment}

In this section, we formulate the spectral assignment problem as a constraint satisfaction problem on a graph, where the goal is to determine a set of expected resonance peaks $q_1,\ldots,q_{m_1}\in \mathbb{R}^3$, from an ensemble of experimentally measured peaks $p_1,\ldots,p_{m_2}\in \mathbb{R}^3$. Each $q_i$ is the frequencies for the coupled triplets associated on each node in Figure~\ref{fig:graph_rep_assignment}. One can consider the list of peaks $ P := \begin{bmatrix}  p_1,\ldots  p_{m_2}\end{bmatrix}\in \mathbb{R}^{3\times {m_2}}$ returned from experiment as a shuffled list of $q_1,\ldots,q_{m_1}$. In the following, we assume ${m_2}\geq {m_1}$, which implies the set of expected resonance peaks is a subset of the set of measured resonance peaks (this is common in practice due to the existence of artifact peaks). This assumption is made to simplify the exposition and is not crucial to the development of the algorithm. The discussion above indicates that the experimental peaks $p_1,\ldots,p_{m_2}$ are related to $q_1,\ldots,q_{m_1}$ via
\begin{equation}
q_i = P z_i\quad i=1,\ldots,{m_1},
\end{equation}
where $z_i\in  \{0,1\}^{m_2}$ and $z_i^T \mathbf{1}_{m_2} = 1$. In other words, $z_i$ is an indicator variable that selects the measured peak from $P$ which in turn provides the values for $q_i$. Therefore determining the indicator variables $z_1,\ldots,z_{m_1}$ is called the spectral assignment problem. When a peak $q_i$ gets assigned a value from one of the $p_1,\ldots,p_{m_2}$, the three atoms that generate peak $q_i$ get assigned with resonance frequencies.

\begin{figure}[]
	\centering
	\includegraphics[width=0.8\columnwidth]{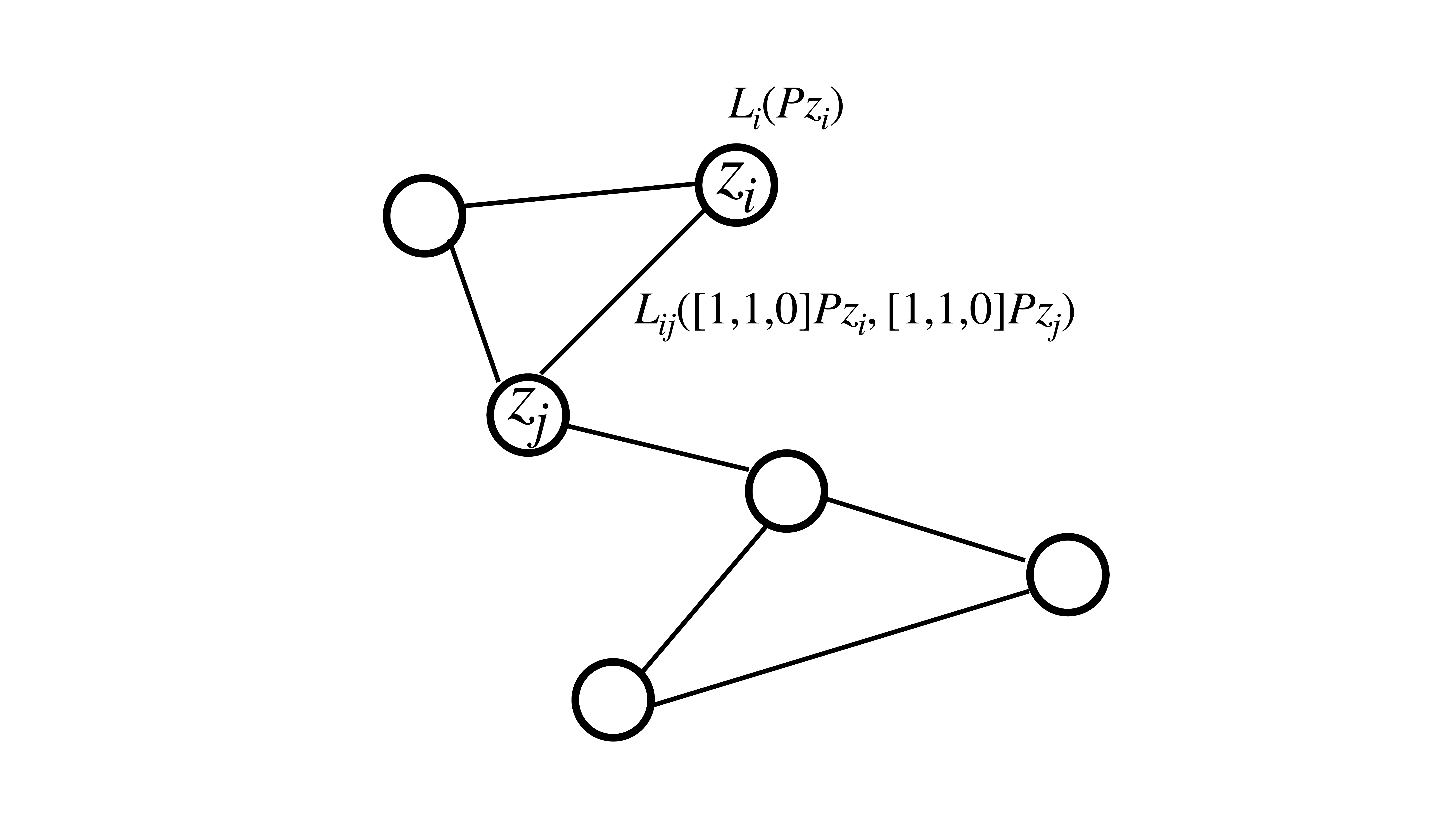}
	\caption{Definition of cost for the assignment problem depicted in Figure~\ref{fig:graph_rep_assignment}. On each edge $(i,j)$ in $G=(V,E)$, there is a loss function $L_{ij}$ penalizing the difference in peaks' frequencies, while on each node a prior term $L_i$ is used to encourage the selection of certain peaks from $P$.}
	\label{fig:graphical_model}
\end{figure}

In order to determine the indicator variables $z_i, i=1,\ldots,{m_1}$, we solve a constraint satisfaction type problem on the graph defined in Figure~\ref{fig:graph_rep_assignment}. Each edge $(i,j)$ is associated with a penalty
\begin{equation}
L_{ij}( B^T_{ij}q_i, B^T_{ij}q_j),
\end{equation}
where each $B_{ij}\in \left\{\begin{bmatrix} 1 & 0 & 0\end{bmatrix}^T, \begin{bmatrix} 0 & 1 & 0\end{bmatrix}^T , \begin{bmatrix} 0 & 0 & 1\end{bmatrix}^T, \begin{bmatrix} 1 & 1 & 0\end{bmatrix}^T, \begin{bmatrix} 0 & 1 & 1\end{bmatrix}^T, \begin{bmatrix} 1 & 0 & 1\end{bmatrix}^T \right\}$, and $L_{ij}$ is some loss function. In words, we want to penalize the difference of certain coordinates between $q_i$ and $q_j$. Furthermore, each node $i$ is associated with a regularization $L_i:\mathbb{R}^3\rightarrow \mathbb{R}$ that is used to impose some prior beliefs on $q_i$. Therefore determining the indicator variables $z_1,\ldots,z_{m_1}$ can be done via solving
\begin{eqnarray}
\min_{\{z_i\}_{i=1}^{m_1}}&\ & \sum_{(i,j)\in E} L_{ij}( B^T_{ij}Pz_i, B^T_{ij}Pz_j) + \sum_{i=1}^{m_1} L_i(P z_i)\label{eq:general assignment}\\
\text{s.t.}&\  &z_i\in  \{0,1\}^{m_2},\ z_i^T \mathbf{1}_{m_2} = 1\ \forall i, \cr
&\ & \sum_{i=1}^{m_1} z_i \leq \mathbf{1}_{m_2},\label{eq:sum_to_1_original}
\end{eqnarray}
an inference problem on a graphical model defined by $G=(V,E)$. The last constraint prevents selecting more than one measured peak from $P$ for each $q_i$. We illustrate this construction in Figure~\ref{fig:graphical_model}.

\begin{figure}[]
	\centering
	\includegraphics[width=0.8\columnwidth]{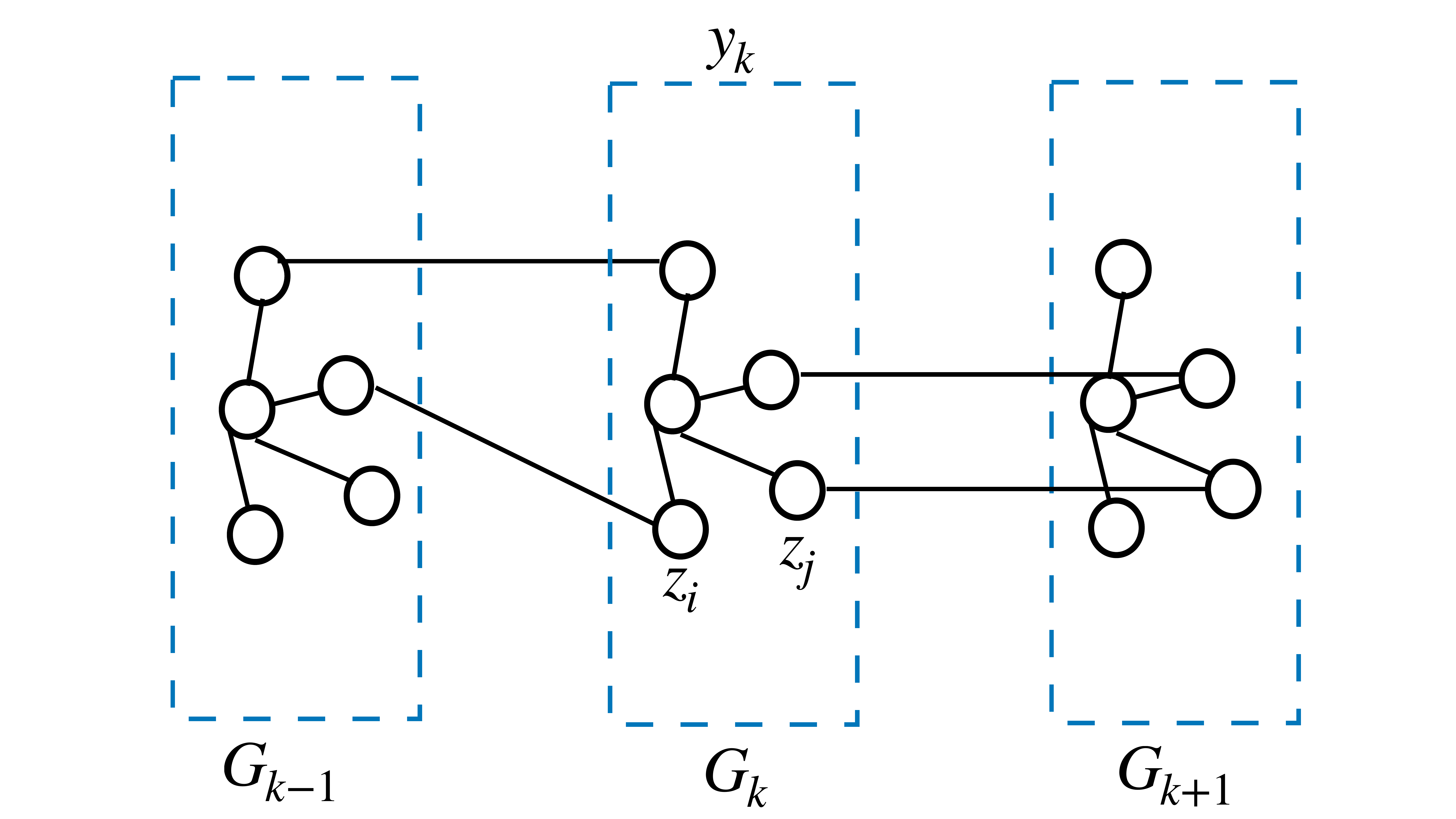}
	\caption{Redefining the variables $z_1,\ldots,z_{m_1}$ in \eqref{eq:general assignment} as $y_1,\ldots,y_n$ in \eqref{eq:general assignment 2} according to subgraph $G_1$,\ldots,$G_n$. Grouping the nodes according to the dotted box induces a path graph.}
	\label{fig:cluster_graphical_model}
\end{figure}

We now turn to a reformulation of \eqref{eq:general assignment} that is closer in spirit to the most common assignment procedure outlined in Section~\ref{section:seq walking}. Suppose there are $n$ residues in the protein. Based on the types of coupling detailed in Section~\ref{section:spectra}, graph $G$ in fact takes the form in Figure~\ref{fig:cluster_graphical_model}, which is a path graph after an appropriate clustering of the nodes into subgraphs $G_1,\ldots,G_n$, where we assume that each subgraph has $c$ nodes. 

Consider unit vectors $y_k, y_{k+1}\in\{0,1\}^{m_2^c}$, each representing the $m_2^c$ ways of associating $c$ measured peaks to each of the node clusters $G_k, G_{k+1}$, respectively (where the dimensionality follows from the fact that there are a total of $m_2$ measured peaks to assign to $c$ nodes). Given some specific choice of $y_k$, denoted by $i$, and some choice of $y_{k+1}$, denoted by $j$, there is some total cost $W_{k, k+1}(i, j)$ associated with the evaluation of the loss functions on our choice of peak assignments. Therefore, we have
\begin{eqnarray}
\min_{\{y_k\}_{i=1}^n}  &\ & \sum_{k=1}^{n-1} \text{Tr}(W_{k,k+1} y_{k+1} y_{k}^T)\label{eq:general assignment 2}\\
\text{s.t.}&\ &y_k\in \{0,1\}^{m_2^c},\ y_k^T \mathbf{1}_{m_2^c}=1,\quad k=1,\ldots,n. \cr
&\ &\mathcal{A}(y_1,\ldots,y_n) \leq \mathbf{1}_{m_2}\label{eq:sum_to_1_cluster} 
\end{eqnarray}
That is, once we group the variables, as depicted in Figure~\ref{fig:cluster_graphical_model}, there are only cost functions defined between the adjacent $y_k,y_{k+1},k=1,\ldots,n-1$. For each possible assignment of measured peaks to each set of adjacent subgraphs there is an associated cost, with the matrix $W_{k, k+1}$ representing the individual costs of all such assignment possibilities. The linear constraint $\mathcal{A}(y_1,\ldots,y_n) \leq \mathbf{1}_{m_2}$ captures \eqref{eq:sum_to_1_original}.

\subsection{Outline of proposed method for solving \eqref{eq:general assignment 2}}
\begin{figure}[]
	\centering
	\includegraphics[width=0.8\columnwidth]{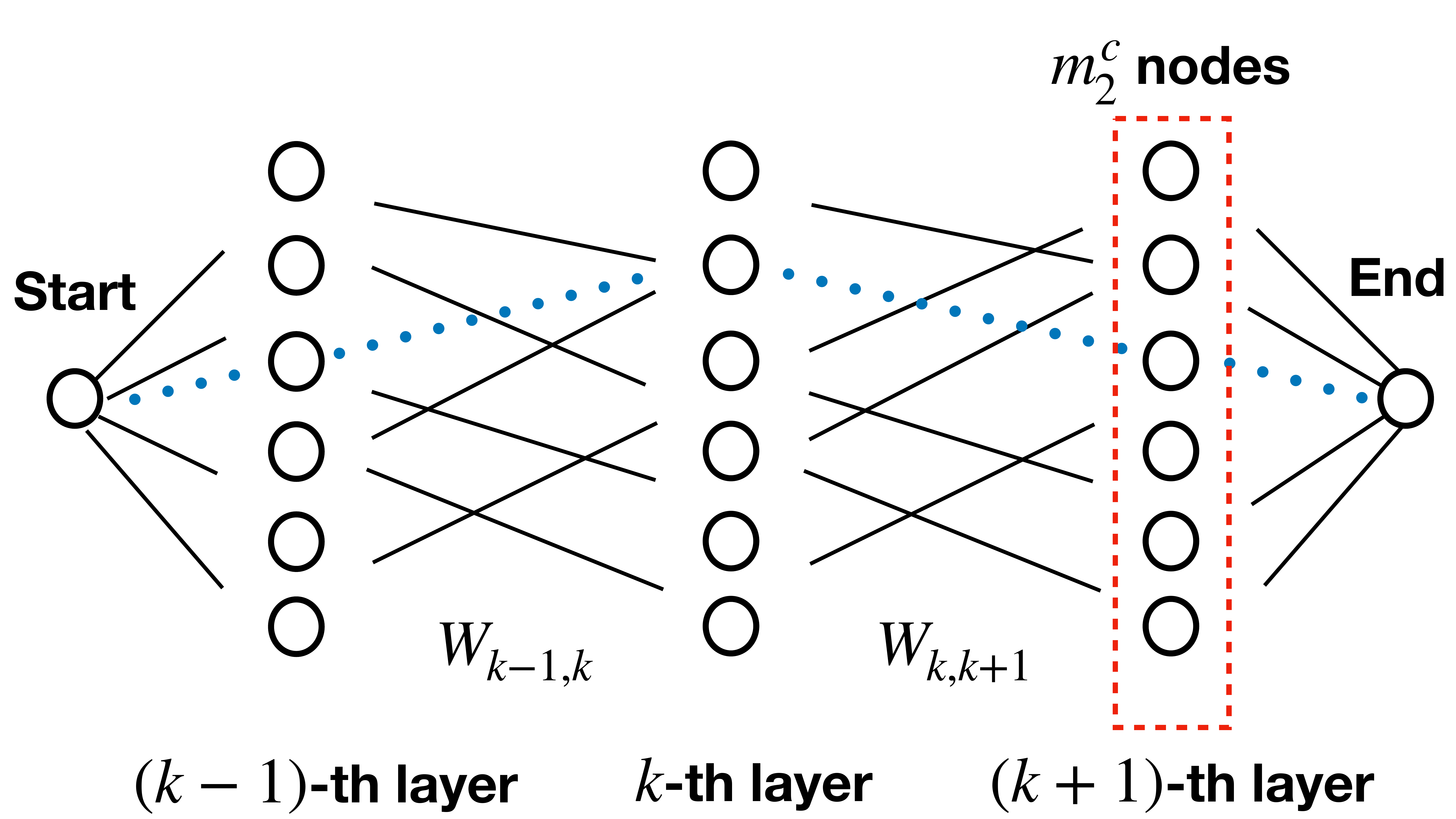}
	\caption{Illustration of finding a solution to \eqref{eq:general assignment 2} by solving a shortest path problem. Here, the shortest path is indicated in blue by dotted line that traverses from the ``start'' to the ``end'' node. Each set of nodes, e.g. nodes in the box, depicts all possible choices for each $y_i$ in \eqref{eq:general assignment 2}. }
	\label{fig:short_path_picture}
\end{figure}

\emph{Without} constraint \eqref{eq:sum_to_1_cluster}, the optimization problem \eqref{eq:general assignment 2} has cost defined on a path graph (since only $y_k$ and $y_{k+1}$, $k=1,\ldots,n-1$ are coupled via some cost functions). This type of optimization problem can be solved using dynamic programming \cite{bang2008digraphs}, and has a complexity of $O(nm_2^{2c})$. More precisely, in order to get a dynamic programming problem, we turn to the construction of a new weighted graph $\mathcal{G} = (\mathcal{V}, \mathcal{E})$ in Figure~\ref{fig:short_path_picture} with $nm_2^{c}+2$ nodes. There are $n+2$ layers, where each layer consists of $m_2^{c}$ nodes (except the first and last layer), and within each layer there are no edges. Edges are formed between two adjacent layers of nodes with weights defined by $\{W_{k,k+1}\}_{k=1}^{n-1}$, where each $W_{k,k+1}\in \mathbb{R}^{m_2^{c}\times m_2^{c}}$. There are two extra nodes in addition to the $nm_2^{c}$ nodes, denoted as the ``start'' and ``end'' nodes. They are connected to the first and last groups of the $nm_2^{c}$ nodes as depicted in Figure~\ref{fig:short_path_picture}. The minimization problem in \eqref{eq:general assignment 2} \emph{without} constraint \eqref{eq:sum_to_1_cluster} thus becomes a problem of tracing the shortest path 
from the start node to the end node, as depicted in Figure~\ref{fig:short_path_picture}. While this problem can be solved efficiently using dynamic programming, such an approach is not possible due to constraint \eqref{eq:sum_to_1_cluster}. Therefore, in the next section, we turn to a linear programming formulation of the shortest path problem, where \eqref{eq:sum_to_1_cluster} can be easily addressed.

\section{Methodology}\label{methodology}
This section presents our general approach to NMR assignment, which we call LIAN (\textbf{LI}near Programming \textbf{A}ssignment for \textbf{N}MR). We first describe how we construct an \emph{assignment graph} $\mathcal{G} = (\mathcal{V},\mathcal{E})$ (different from $G=(V,E)$ in \eqref{eq:general graphical MLE} or \eqref{eq:general assignment}) on which we efficiently solve \eqref{eq:general assignment 2}, a constrained shortest path problem whose solution yields a valid assignment which approximately maximizes the expected log-likelihood under a given probabilistic model. We divide this section into two parts: (1) defining the structure of the assignment graph (Section~\ref{ch-lp-graph}), and (2) solving the constrained shortest path problem (Section~\ref{sec:shortest path}). 
\subsection{Building the assignment graph\label{ch-lp-graph}}

The assignment graph $\mathcal{G}$ is a directed graph with $n+2$ layers, where $n$ is the known number of residues in the protein. This is illustrated in Figure~\ref{fig:short_path_picture}, where a layer is a group of nodes, for example those in the red dotted box. The construction of the graph proceeds in three steps:

\begin{enumerate}
    \item \textbf{Initial peak groupings}: We first enumerate groups of measured peaks whose frequencies are internally consistent.
    More precisely, we partition the variables $z_1,\ldots,z_{m_1}$ in \eqref{eq:general assignment} associated with nodes of $G$, by partitioning $G$ into $n$ subgraphs $G_1,\ldots, G_n$ (corresponding to $n$ residues) as in Figure~\ref{fig:cluster_graphical_model}, and for each part we enumerate all the possible choices of assignments. For example, if each $G_k$ has $c$ number of $z_i$'s associated with it, and each $z_i$ has $m_2$ choices, then there are at most $m_2^c$ choices for all the variables in $G_k$. This is too large in general, therefore we pick the possible values for all $z_i, z_j$ associated with $G_k$ such that $L_{ij}(B_{ij}^TPz_i, B_{ij}^T P z_j)$ is smaller than some threshold, for all $i,j\in G_k$. The possible values of $c$ $z_i$'s within each part essentially corresponds to a choice of $c$ peaks from $p_1,\ldots,p_{m_2}$. Therefore this is called the peak grouping procedure. We assume there are $g$ choices for the variables in each $G_k$, which give $g$ nodes in the $k$-th layer of the assignment graph.
	\item \textbf{Creating the graph nodes}: A possible combination of peaks in $P$, that can be assigned to the nodes in $G_k$, forms a node in a $k$-layer of $\mathcal{G}$. Again, corresponding to each layer (i.e. each residue), there are $g$ nodes. To further cut down the number of nodes, for each residue we enumerate the peak groupings that are sufficiently consistent with each residue, as determined by the difference between the frequencies in the peak grouping and a prior, which is derived from chemical shift statistics for each residue type stored in BMRB \cite{ass:BMRB}. This step associates a cost related to the log-likelihood of the assignment under the prior for each node. For the $k$-th residue, such a cost basically comes from $L_{i}(Pz_i), i\in G_k$ in \eqref{eq:general assignment}. After this step, the $k$-th layer is left with $g_k$ nodes.
	\item \textbf{Creating the graph edges}: Edges \emph{between} two layers are added to the graph between any two nodes which have sufficiently consistent frequency assignments for the same atoms. Each edge contributes a cost commensurate with the relevant level of consistency. Such a cost comes from $L_{ij}(B_ij^TPz_i, B_{ij}^T P z_j), i\in G_k, j\in G_{k+1}$ in \eqref{eq:general assignment}.
\end{enumerate}

We describe each of these steps in greater detail below.

\subsubsection{Initial peak groupings}\label{sec:peak_groupings}
As explained in Section \ref{section:assignment}, our experimental data consists of a list of peaks, $P:=[p_1,\ldots, p_{m}]$, where each $p_i\in \mathbb{R}^{3}$ corresponds to a set of atom frequencies. In order to form nodes from this list of peaks, we group them in groups that are internally consistent, i.e., groups of peaks which assign approximately the same frequency to the same atom.

As mentioned previously, a protein consists of $n$ residues that have repeated sets of atoms. The $k$-th residue $r_k$ contains atoms $\aN_k$, $\aH^N_k$, $\aC_k^\alpha$, $\aC_k^\beta$. In subgraph $G_k$, the nodes come from residues $r_k$ and $r_{k+1}$ with triplets forming by $\aN_k$, $\aH^N_k$, $\aC_k^\alpha$, $\aC_k^\beta$,  $\aC_{k+1}^\alpha$, $\aC_{k+1}^\beta$. When considering an NMR dataset with three spectra, HSQC, HNCACB, and HN(CO)CACB, we expect $G_k$ to contain seven nodes, coming from the fact that there are seven triplet interactions all involving the same $\aN_k$ and $\aH^N_k$. Therefore, each $G_k$ contributes to seven peaks in the three spectra. Some of these peaks will also share $\aC_k^\alpha$ and $\aC_k^\beta$ frequencies. This is illustrated in Figure~\ref{fig:3-spectra}, where all seven peaks have consistent frequency values in \fifteenn--\oneh{N} plane (and there are two peaks agreeing along the $\aC$ dimension). This allows us to guess valid peak groupings associated with $G_k$. To this end, we make use of an enumeration procedure (described in Appendix \ref{app:graph_construction}) to enumerate all consistent peak groupings, which are defined as groupings of seven peaks (or more, depending on the experiment set)  where the frequencies associated with certain atoms do not differ by more than an experimentally-accepted threshold. We describe it more concisely here in the context of our example:

\begin{enumerate}
    \item Select a reference spectrum (often called a \emph{fingerprint} spectrum). This is typically a spectrum from an experiment such as HSQC, which contains the peaks generated from the pairs $(\aN_k, \aH^N_k)$ as these spectra have higher sensitivity than other experiments and are therefore less likely to be missing peaks. In principle, there should be $n$ peaks in this spectra. 
    \item For each HSQC peak, enumerate all peaks in other spectra which are consistent with peaks in the fingerprint spectrum along the \fifteenn--\oneh{N} dimensions, within appropriate experimental thresholds $(\delta_1, \delta_2)$.
    \item Among all consistent peaks, identify all subsets which are consistent (within experimental threshold $\delta_3$) along the corresponding $\aC$ dimension.
\end{enumerate}

\paragraph{Spin systems:} On some occasions, NMR practitioners perform this grouping procedure manually (or in a human-in-the-loop, computer-guided fashion). The groupings of measured peaks are then summarized in the form of \emph{spin systems}, by averaging the frequencies assigned to each atom, thus producing a simple vector of "consensus" atom frequencies. As a result, data is sometimes summarized in the spin system format rather as peak lists, and we can skip the grouping step described in this section. However, we note that this leads to some information loss, as we lose information related to the level of agreement between frequencies assigned to the same atom by different peaks.

\subsubsection{Creating the graph nodes}\label{sec:graph_nodes}
Nodes in the assignment graph are subdivided into $n+2$ layers, one for each of the $n$ residues in the protein, and additional \emph{start} and \emph{end} layers to simplify the formulation of the problem. There are three broad classes of nodes:
\begin{enumerate}
    \item \textbf{Start and End nodes}: The first layer and the $n+2$-th layer consist of a single node, used for convenience. These nodes help define the start and end position of the shortest path we seek to find.
    \item \textbf{Dummy nodes}: There is one such node for each of the $n$ inner layers, and their function is to ensure the shortest path problem is feasible. There is a path from every node in layer $k-1$ to the dummy node of layer $k$, and from this dummy node to every node in layer $k+1$. If included in the final path, no frequencies will be assigned to the atoms in residue $k$, such that this node is equivalent to a \emph{null} assignment for residue $k$, and incurs a high associated cost.
    \item \textbf{Regular nodes}: All other nodes in the graph represent a grouping of measured peaks, as defined in Section \ref{sec:peak_groupings}, which is consistent with the given residue. 
\end{enumerate}

In sum, there is exactly one start and one end node, and there are exactly $n$ dummy nodes, one for each residue of the protein. However, given $g$ valid peak groupings (as defined in Section \ref{sec:peak_groupings}) we create $g_k\leq g$ nodes in each layer. This is because any peak groupings which are not consistent with the prior on the atom frequencies for a given residue are not instantiated, in order to reduce the overall size of the graph. We formalize the process for eliminating nodes below, upon introducing the edge cost definitions.

\subsubsection{Creating the graph edges}

After creating all $n+2$ node layers, we connect nodes between each layer and the subsequent layer. The edge creation step is most important because it is also where we define the costs associated with each edge. In sum, we want this cost to represent some notion of probabilistic agreement between our assumed generative model for the data and our set of observations. 

\begin{figure}
\centering
    \begin{tikzpicture}
    
    
      \node[obs, xshift=-2cm] (x1)   {$x_1^a$}; %
      \factor[above=of x1] {x1-f1} {left:$\mathcal{N}$} {} {} ; %
      \node[const, right=0.4cm of x1-f1] (s1) {$\sigma_1$}; %
      
      \node[obs] (x2)   {$\ldots$}; %
      \factor[above=of x2] {x2-f2} {left:$\mathcal{N}$} {} {} ; %
      \node[const, right=0.4cm of x2-f2] (s2) {$\sigma_{\ldots}$}; %
      
      \node[obs, xshift=2cm] (x3)   {$x_{o_a}^a$}; %
      \factor[above=of x3] {x3-f3} {left:$\mathcal{N}$} {} {} ; %
      \node[const, right=0.4cm of x3-f3] (s3) {$\sigma_{o_a}$}; %
      
      \node[latent, above=of x2-f2] (mu)   {$\mu$}; %
      \factor[above=of mu] {mu-fmu} {left:$\mathcal{N}$} {} {} ; %
      
      \node[const, above=1.2 of mu, xshift=-0.5cm] (muw) {$\mu_a$}; %
      \node[const, above=1.2 of mu, xshift=0.5cm] (mus) {$\sigma_a$}; %
      
      \factoredge {muw, mus} {mu-fmu} {mu}; %
      \factoredge {s1, mu} {x1-f1} {x1}; %
      \factoredge {s2, mu} {x2-f2} {x2}; %
      \factoredge {s3, mu} {x3-f3} {x3}; %
      
    \end{tikzpicture}
    \caption{Generative model for an atom observed by $o_a$ peaks. A Gaussian prior for the frequency of each atom, $a$, is assumed, with parameters $\mu_a$ and $\sigma_a$ derived from chemical shift statistics deposited in BMRB \cite{ass:BMRB}. The observed frequencies  $\{x^a_1, \ldots, x^a_{o_a}\}$ of each atom are also assumed to be normally distributed, centered around the latent frequency of the atom and with (assumed known) experimental variance $\sigma_1,\ldots,\sigma_{o_a}$.}
    \label{fig:gen-model}
\end{figure}
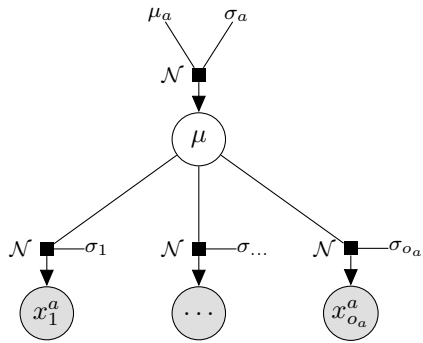

\paragraph{Generative model:} In Figure~\ref{fig:graph_rep_assignment}, atom $a_1$ is shared by two nodes in graph $G$. That means two peaks observed in the spectra are associated with $a_1$. We want to model the probability distribution of the observed peaks associated with an atom in common. Many probabilistic cost functions would be reasonable, but for the purposes of this paper we assume that the prior on each atom's frequency is Gaussian, and that the experimental noise is also Gaussian (with mean 0). This is depicted in Figure \ref{fig:gen-model} for an atom, $a$, for which we have $o_a$ distinct observations of its frequency, $\{x^a_1, \ldots, x^a_{o_a}\}$. This implies that this atom is associated with $o_a$ peaks (or in other words associated with $o_a$ nodes in graph $G$ in Figure~\ref{fig:graph_rep_assignment}). Such a generative model prescribes a graphical model on graph $G$. Solving \eqref{eq:general assignment} amounts to performing inference on $z_i$'s under such a probabilistic model. This generative model is consistent with much of the automated assignment literature (see, e.g. \cite{Cisa}) with the notable exception of FLYA, which assumes the experimental noise is a truncated Gaussian, in order to guarantee feasible assignments under its definition of a valid assignment \cite{ass:flya}. 

Under this model, we define the score associated with each atom as
\begin{defn}[Atom cost]\label{def:atom cost}
	The cost associated with atom $a$, with a normally distributed prior $\mcal{N}(\mu_a, \sigma_a)$, and $o_a$ observations $\{x_l^a\}_{l=1}^{o_a}$ defined by the peak grouping, also assumed to be normally distributed around the true frequency, $\mu$, according to $\mcal{N}(\mu, \sigma_l)$ is defined as
	\begin{equation}\label{eq:atom cost}
	\text{cost}\left(a, \{x_l^a\}_{l=1}^{o_a}\right) \triangleq -\log\mathbb{E}_{\mu\sim \mathcal{N}(\mu_a, \sigma_a)}\left[\prod_{l=1}^{o_a}f(x_l^a\mid \mu, \sigma_l)\right].
	\end{equation}
	where $f(\cdot\mid u, v)$ is the Gaussian density with mean $u$ and standard deviation $v$. This expectation works out to a simple expression involving the observations, experimental noise parameters, and the parameters of the prior distribution, as explained in Appendix \ref{app:graph_construction}.
\end{defn}

Now, recall that if we select an edge between two nodes, node $i$ in layer $k$, and node $j$ in layer $k+1$ in Figure~\ref{fig:short_path_picture}, to be included in our path, we are indeed assigning observed peaks to the nodes in $G_k$ and $G_{k+1}$. This implies that all the atoms involved in establishing the nodes (recall that each node is associated with a triplet or a pair of atoms) are assigned a frequency valued obtained from the observed peaks. Then the generative model in Definition~\ref{def:atom cost} determines how likely these frequency assignments are under the assumed generative model, which will, in turn, help determine the likelihood of an edge $(i, j)$ in Figure~\ref{fig:short_path_picture}.

This provides a cost for selecting edge between the $k$-th and $(k+1)$-th layers. Let $r_{k+1}$ be the set of backbone atoms associated with residue $k+1$. The peak groupings in the two nodes connected by edge $(i, j)$ in $\mathcal{G}=(\mathcal{V},\mathcal{E})$ imply the assignment of a set of observations $\{x_l^a\}_{l=1}^{o_a}$ for each atom in the set of backbone atoms $r_{k+1}$. Since only nodes in layers $k$ and $k+1$ include observations for the atoms in $r_{k+1}$, we define the cost of the edge as follows:

\begin{defn}[Edge cost]\label{def:edge-cost}
	Each edge between node $i$ in layer $k$ and node $j$ in layer $k+1$ in in $\mathcal{G} = (\mathcal{V},\mathcal{E})$ assigned frequencies $\{x_l^a\}_{l=1}^{o_a}$ to each atom $a\in r_{k+1}$.  The edge cost is then defined as
	\begin{equation}\label{eq:edge-cost}
	\text{edge cost}\left(r_{k+1}, i, j\right) \triangleq \sum_{a\in r_{k+1}} \text{cost}\left(a, \{x_l^a\}_{l=1}^{o_a}\right).
	\end{equation}
\end{defn}

As such, each edge between layers $k$ and $k+1$ incorporates the cost associated with \emph{all} observations on the atoms in residue $k+1$ induced by the peak groupings in the relevant nodes.

\subsubsection{Statistical Typing}
In order to further manage the size of the assignment graph, edge whose associated cost is too large (representing an extremely unlikely assignment) can be discarded at this stage. These are typically edges between nodes whose induced frequencies disagree strongly with the prior distributions for a residue's atoms. Details about how we set the threshold for inclusion can be found in Appendix \ref{app:statistical-typing}.\\

\subsection{Finding a shortest path in a directed graph}\label{sec:shortest path}

Having constructed the assignment graph, we formulate the assignment problem as one of finding a shortest path in a directed graph, $\mcal{G}=(\mcal{V}, \mcal{E})$, subject to some utilization constraints. The cost of any path from the start to the end node is equal to the expected negative log-likelihood of the assignment induced by that path. As a result, finding the shortest path in this graph amounts of finding the path of greatest expected log-likelihood. We highlight that alternative formulations of the cost are also possible (such as, for example, directly maximizing the log-likelihood, rather than the expected likelihood). However, the formulation presented here is computationally straightforward to implement, allowing us to quickly build large graphs.

While an unconstrained shortest path problem is straightforward to solve through dynamic programming, our problem is not unconstrained, due to the fact that each observed datum (i.e. a measured peak) can only be utilized once in a valid assignment. This practical limitation can be concisely written as a linear constraint, which fits into the constraint satisfaction framework we describe in Section~\ref{section:assignment}. 

In particular, recall that we had defined the assignment problem \eqref{eq:general assignment 2}, where $y_k\in \{0, 1\}^{m_2^c}$ is a unit vector selecting a group of peaks for $G_k$. After reducing the possible choices, we actually redefine our $y_k$ variables into $y_k\in\{0, 1\}^{g_k}$, assuming $g_k$ valid combinations for layer $k$. We can further define
\begin{equation}
    X_{k, k+1}\triangleq y_{k}y_{k+1}^T \in \{0, 1\}^{g_k\times g_{k+1}}, \qquad \mbf{1}_{g_k}^TX_{k, k+1}\mbf{1}_{g_{k+1}}=1
\end{equation}
as a selection matrix, where $X_{k, k+1}(i, j)=1$ implies that node $i$ is selected in layer $k$ and node $j$ is selected in layer $k+1$. Under this simple redefinition, $W_{k, k+1}\in\mathbb{R}^{g_k\times g_{k+1}}$ can also be easily understood as the matrix of edge costs between layers $k$ and $k+1$. That is
\begin{equation}\label{eq:wk}
    W_{k, k+1}(i, j)\triangleq \text{edge cost}(i, j)
\end{equation}
where $\text{edge cost}(i, j)$ is the cost associated with the edge between node $i$ in layer $k$ and node $j$ in layer $k+1$. Also note that we need not worry about layers $0$ and $n+1$ to express the linear programming formulation of the problem, since there is a single node in these two layers. 

Finally, we can formulate the NMR assignment problem in terms of these new variables:
\begin{prob}[NMR assignment]\label{lp-prob-r0}
    \begin{eqnarray}
    \min_{\{X_{k, k+1}\}_{k=0}^{n}}  &\ & \sum_{k=1}^{n} \text{Tr}(W_{k,k+1}^T X_{k, k+1})\\
    \text{s.t.}&\ &X_{k, k+1}\in  \{0, 1\}^{g_k\times g_{k+1}},\ \mbf{1}_{g_k}^TX_{k, k+1}\mbf{1}_{g_{k+1}}=1, k=1,\ldots,n \cr
    &\ &X_{k-1, k}^T\mbf{1}_{g_{k-1}}=X_{k, k+1}\mbf{1}_{g_{k+1}}, k=1, \ldots, n \cr
    &\ &{\mathcal{A}}(X_{1, 2}\mbf{1}_{g_2}, \ldots, X_{n, n+1}\mbf{1}_{g_{n+1}}) \leq \mathbf{1}_{m_2}
    \end{eqnarray}
\end{prob}

Note the following details: A \emph{path} constraint is included in the formulation, enforcing that the end node selected using $X_{k-1, k}$ must coincide with the start node selected using $X_{k, k+1}$. Compare to \eqref{eq:general assignment 2}, the summation in the cost of \eqref{lp-prob-r0} goes from $k=0$ to $k=n$, which takes into account of the extra ``start'' and ``end'' node, and $g_0=g_{n+1}=1$.

The problem as formulated above is equivalent to a constrained shortest path problem, and is NP-hard \cite{NetFlows}. For small enough problems, integer linear programming (ILP) solvers such as Gurobi \cite{gurobi} can successfully solve the problem with short runtimes. In our experience, this is often feasible whenever the input consists of a high quality set of nodes in each layer of $\mathcal{G}$ (e.g. when we have a set of reliable spin systems as input). However, for larger problems we can instead make use of linear programming relaxations of Problem \ref{lp-prob-r0}. These relaxations can occasionally return integer solutions (in which case the solution coincides with the solution for Problem \ref{lp-prob-r0}) or, more commonly, partially-integer solutions (i.e. a solution in which many of the entries in each $X_{k, k+1}$ are integer, with most being zero), from which one can then derive a satisfactory solution as we describe shortly below. We make use of the following relaxation:

\begin{prob}[NMR assignment, LIAN-1]\label{lp-prob-r1}
    \begin{eqnarray}
    \min_{\{X_{k, k+1}\}_{k=0}^{n}}  &\ & \sum_{k=0}^{n} \text{Tr}(W_{k,k+1}^T X_{k, k+1})\\
    \text{s.t.}&\ &X_{k, k+1}\geq 0, X_{k, k+1}\leq 1, k=1, \ldots, n\cr
    &\ &\mbf{1}_{g_k}^TX_{k, k+1}\mbf{1}_{g_{k+1}}=1, k=1,\ldots,n \cr
    &\ &X_{k-1, k}^T\mbf{1}_{g_{k-1}}=X_{k, k+1}\mbf{1}_{g_{k+1}}, k=1, \ldots, n \cr
    &\ &{\mathcal{A}}(X_{1, 2}\mbf{1}_{g_2}, \ldots, X_{n, n+1}\mbf{1}_{g_{n+1}}) \leq \mathbf{1}_{m_2}\label{eq:sum_to_1_cluster-2} 
    \end{eqnarray}
\end{prob}
This follows from relaxing the original (matrix-integer) variables $X_{k, k+1}$ to their convex hull, where $X_{k, k+1}\in [0, 1]^{g_k\times g_{k+1}}$ and $\mbf{1}_{g_k}^TX_{k, k+1}\mbf{1}_{g_{k+1}}=1$. As alluded to above, solving this problem typically results in a sparse, partially-integer solution. That is: a solution in which some of the entries in each $X^*_{k, k+1}$ that solves Problem \ref{lp-prob-r1} are integer, with most being zero. Such a solution induces a much smaller subgraph of the original assignment graph by retaining only edges, $(i, j)$, for which $X^*_{k, k+1}(i, j)\neq 0$. We then solve Problem \ref{lp-prob-r0} on that induced subgraph.

We note that the utilization constraint \eqref{eq:sum_to_1_cluster-2} can sometimes be too strict as there are often peaks which are overlapping, resulting less peaks than expected. As a result, strictly preventing data from being re-utilized can hurt, rather than help, the solution. To address this issue, we also consider an alternative relaxation, as follows:

\begin{prob}[NMR assignment, LIAN-2]\label{lp-prob-r2}
    \begin{eqnarray}
    \min_{\{X_{k, k+1}\}_{k=0}^{n}}  &\ & \sum_{k=0}^{n} \text{Tr}(W_{k,k+1}^T X_{k, k+1}) + \lambda \mbf{1}_{m_2}^T \epsilon\label{eq:nmr-assignment-1}\\
    \text{s.t.}&\ &X_{k, k+1}\geq 0, X_{k, k+1}\leq 1, k=1, \ldots, n\cr
    &\ &\mbf{1}_{g_k}^TX_{k, k+1}\mbf{1}_{g_{k+1}}=1, k=1,\ldots,n \cr
    &\ &X_{k-1, k}^T\mbf{1}_{g_{k-1}}=X_{k, k+1}\mbf{1}_{g_{k+1}}, k=1, \ldots, n \label{eq:path_constrain}\cr
    &\ &{\mathcal{A}}(X_{1, 2}\mbf{1}_{g_2}, \ldots, X_{n, n+1}\mbf{1}_{g_{n+1}})-\epsilon = \mathbf{1}_{m_2}\cr
    &\ &\epsilon \geq 0, \epsilon \in\mathbb{R}^{m_2}
    \end{eqnarray}
\end{prob}

Note that this relaxation penalizes (but allows for) the reutilization of measured peaks/spin systems at multiple points of the assignment through the use of slack variable $\epsilon$. The reutilization of peaks is penalized in the cost function, with each reutilization costing $\lambda$ in added cost. This $\lambda$ thus becomes a user-set parameter.

\begin{figure}[!h]
	\centering
	\includegraphics[width=\columnwidth]{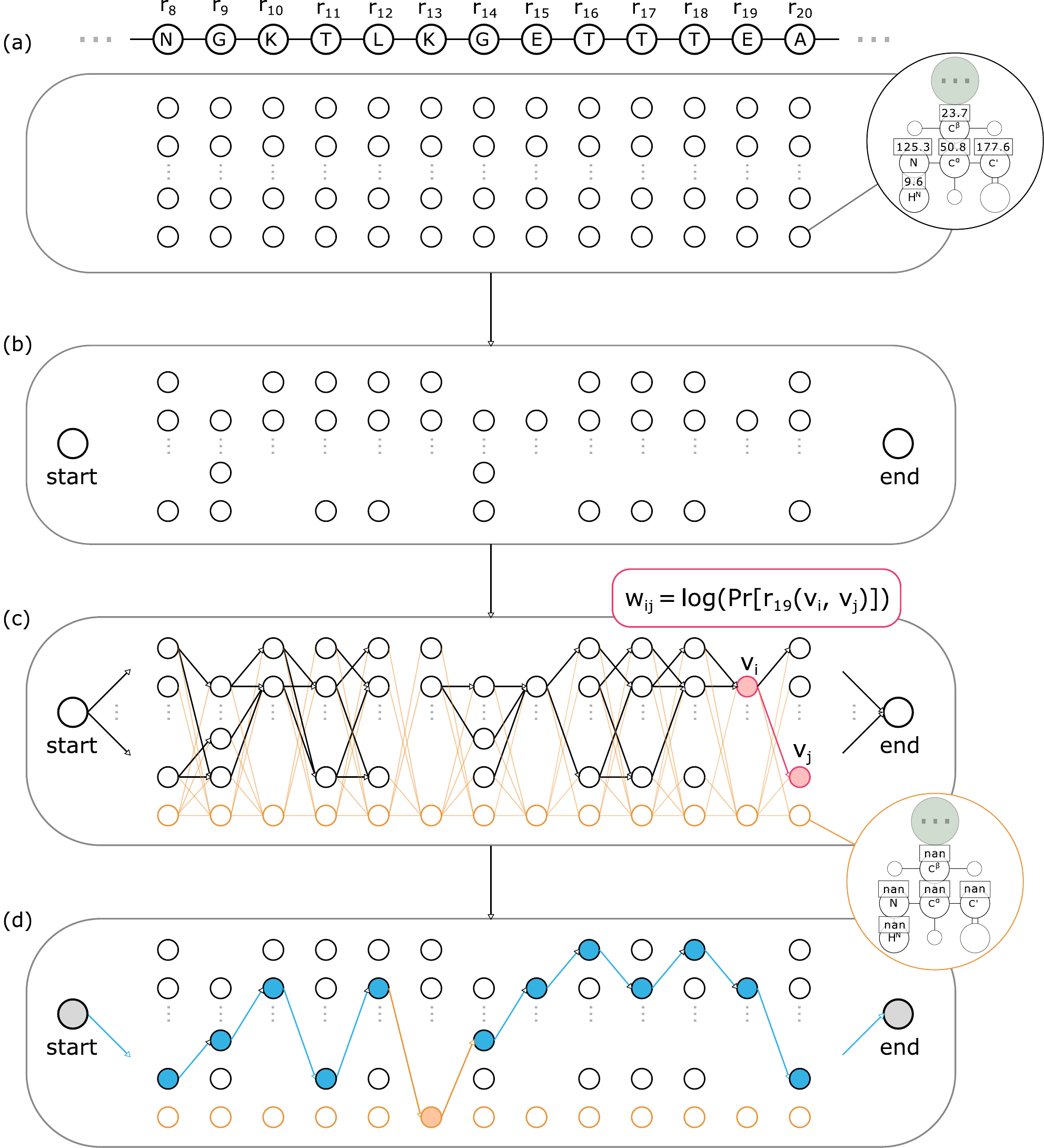}
	\caption[Illustration of graph assignment model.]{Illustration of the full assignment procedure. (a) Possible chemical shift assignments are determined and enumerated for each residue, creating nodes in a graph. (b) Each node is statistically typed against its residue's distribution, and very low likelihood nodes are eliminated. (c) Edges are placed between nodes $i$, $j$ in adjacent layers, $k$ and $k+1$ with weight equal to the posterior log-likelihood of the respective assignment. Empty nodes are added to each residue and connected to every node in the preceding and succeeding layers with edge weights equal to the threshold. (d) A longest path is found between the start and end nodes, subject to any additional constraints (e.g. that spin systems cannot be used more than once).}
	\label{fig:lp-model}
\end{figure}

An end-to-end description of the full assignment procedure is summarized in Figure \ref{fig:lp-model}.

\subsection{Notes on the methodology}
We emphasize that the graph-based approach presented in this paper extends to any objective function that can be expressed in the form of Problem \ref{lp-prob-r0}. As a result, many different generative models can easily be accommodated by this methodology, and one is also not limited to maximizing an expected log-likelihood. As an example, for each edge in the assignment graph connecting two nodes between layers $k$ and $k+1$, which assign a set of observed chemical shifts to the atoms of residue $k$, one could solve a local maximum likelihood problem for that residue under the given observations. Doing so would enable a maximum likelihood estimate instead.

We also note that prior information or information from supporting experiments can also be accommodated, either by modifying the edge costs (through the matrices $W_{k, k+1}$) or by editing the assignment graph. As a concrete example, if one had accurate information about the chemical shifts in a particular residue, $r_{k+1}$, then one could specify tight priors, $\mathcal{N}(\mu_a, \sigma_a)$, for each atom $a$ in that residue, where $\mu_a$ would be set to the chemical shift implied by the prior information, and $\sigma_a$ would be set to a value that accurately represents the uncertainty of that prior information. By doing so, Equations \ref{eq:atom cost}, \ref{eq:edge-cost}, and \ref{eq:wk} would result in entries of $W_{k, k+1}$ which are high for any chemical shift assignment that violates the prior information about residue $r_{k+1}$, and low for any assignment consistent with that prior information, as desired. Additionally, one could also use this prior information to directly edit the assignment graph (i.e. by removing any nodes which represent an assignment inconsistent with prior knowledge). This not only helps enforce prior knowledge, but also helps the efficiency of the algorithm.

In sum, while we describe a concrete end-to-end methodology using a specific generative model, the building blocks of our approach should prove more broadly applicable to a wider range of problems in spectral assignment.
\section{Results}\label{results}

\subsection{Simulated data}

\subsubsection{CISA}

As a first sanity check, we tested our approach on the entirety of the benchmark dataset developed by the authors of CISA in \cite{Cisa}, as it provides a useful comparison to many other fully automated algorithms on problems of small, medium, and large scale. This synthetic dataset is created by generating a simulated list of spin systems from the ground-truth assignment values recorded in BMRB \cite{ass:BMRB}. White Gaussian noise is then added to the carbon atoms, with standard deviations $\sigma_\alpha=0.08$ ppm, $\sigma_\beta=0.16$ ppm for the \thirteenc{\alpha} and \thirteenc{\beta} atoms, respectively, in the \emph{low-noise} simulation, and $\sigma_\alpha=0.16$ ppm, $\sigma_\beta=0.32$ ppm in the \emph{high-noise} simulation. For full details on the simulation scenario, we refer the reader to \cite{Cisa}.

We compare the performance of LIAN to that of 4 different algorithms. Results for MARS \cite{Mars} and CISA \cite{Cisa} are both retrieved from the original CISA paper. We also compare with IPASS \cite{ass:ipass}, although we note that the relevant paper does not mention adding noise to the spin systems (and, indeed, refers to a "perfect connectivity" scenario, which suggests that no noise is added). Finally, we also include a partial comparison with C-SDP \cite{Ferreira2018}, which is an earlier semidefinite programming relaxation approach to the NMR assignment problem which we developed, but which does not scale to larger proteins. The results for LIAN were obtained by solving Problem \ref{lp-prob-r1}, which typically produces a partially integer solution. This partially integer solution induces a (much) smaller subgraph on which we can efficiently solve the original integer linear programming problem. Each row is averaged over 100 simulations.

The results are summarized in Tables \ref{tab:synthetic_low_lp} and \ref{tab:synthetic_high_lp}, for the two distinct noise levels considered in \cite{Cisa}. We evaluate the results by calculating the precision and recall of the assignment algorithm. Let $m_\text{assigned}$ be the number of assigned residues (i.e. non-dummy nodes in the path), $m_\text{correct}$ be the number of \emph{correctly} assigned residues, and $m_\text{assignable}$ be the number of assignable residues (i.e. residues which have a ground-truth assignment). Then
\begin{align*}
    \text{precision}&\triangleq m_\text{correct}/m_\text{assigned} \\
    \text{recall}&\triangleq m_\text{correct}/m_\text{assignable}.
\end{align*} 

\begin{table}[h!]\footnotesize
	\centering
	\caption[Performance of LIAN-1 on simulated spin datasets (low noise)]{ Accuracy of assignment (precision/recall) of various algorithms and LP on synthetic spin systems with noise level ($\sigma_\alpha$, $\sigma_\beta$) $= (0.08, 0.16)$. Results for MARS \cite{Mars} and CISA taken from Table 2 in \cite{Cisa}. Results for IPASS taken from Table 3 in \cite{ass:ipass}. Results for C-SDP taken from Table 1 in \cite{Ferreira2018}.}
	\label{tab:synthetic_low_lp}
	\begin{tabular}{cccccccc}
		\hline\\
		\textbf{Protein ID} & \textbf{Length} & \textbf{N}${}^1$ & \textbf{MARS} & \textbf{CISA} & \textbf{IPASS} & \textbf{C-SDP} & \textbf{LIAN-1}\\ \hline
		bmr4391 & 66 & 59 & 91/97 & 97/97 & 93/90 & 99/99 & 90/90 \\
		bmr4752 & 68 & 66 & 98/97 & 96/94 & 100/94 & 100/100 & 100/100 \\
		bmr4144 & 78 & 68 & 100/97 & 100/99 & 98/85 & 100/100 & 99/96     \\
		bmr4579 & 86 & 83 & 97/91 & 98/98 & 100/98 & 100/100 & 100/99 \\
		bmr4316 & 89 & 85 & 97/96 & 100/99 & 99/98 & 99/99 & 100/100\\
		bmr4288 & 105 & 94 & 97/95 & 98/98 & 100/98 & & 99/99 \\
		bmr4929 & 114 & 110 & 99/97 & 93/91 & 100/100 & & 100/98 \\
		bmr4302 & 115 & 107 & 95/92 & 96/95 & 100/99 & & 100/99 \\
		bmr4670 & 120 & 102 & 94/88 & 96/95 & 98/97 & & 99/99 \\
		bmr4353 & 126 & 98 & 91/85 & 96/95 & 99/93 & & 95/95 \\
		bmr4207 & 158 & 148 & 96/93 & 100/99 & 100/97 & & 99/99\\
		bmr4318 & 215 & 191 & 88/81 & 87/84 & 100/98 & & 98/98\\
		\hline 
	\end{tabular}
	\flushleft
	{\footnotesize ${}^1$ Number of assignable spin systems in the BMRB data.}
\end{table}

\begin{table}[h!]\footnotesize
	\centering
	\caption[Performance of LIAN-1 on simulated spin datasets (high noise)]{Accuracy of assignment (precision/recall) of various algorithms and LP on synthetic spin systems with noise level ($\sigma_\alpha$, $\sigma_\beta$) $= (0.16, 0.32)$. Results for MARS \cite{Mars} and CISA taken from Table 2 in \cite{Cisa}. Results for IPASS taken from Table 3 in \cite{ass:ipass}. Results for C-SDP taken from Table 2 in \cite{Ferreira2018}. }
	\label{tab:synthetic_high_lp}
	\begin{tabular}{cccccccc}
		\hline\\
		\textbf{Protein ID} & \textbf{Length} & \textbf{N}${}^1$ & \textbf{MARS} & \textbf{CISA} & \textbf{IPASS} & \textbf{C-SDP} & \textbf{LIAN-1}\\ \hline
		bmr4391 & 66 & 59 & 86/85 & 91/91 & 93/90 & 100/100 & 86/86 \\
		bmr4752 & 68 & 66 & 91/90 & 90/88 & 100/94 & 99/99 & 100/100 \\
		bmr4144 & 78 & 68 & 100/97 & 100/99 & 98/85 & 96/96 & 96/94 \\
		bmr4579 & 86 & 83 & 79/75 & 80/80 & 100/98 & 100/100 & 100/99 \\
		bmr4316 & 89 & 85 & 95/92 & 83/83 & 99/98 & 98/98 & 99/99 \\
		bmr4288 & 105 & 94 & 95/93 & 91/91 & 100/98 & & 99/99 \\
		bmr4929 & 114 & 110 & 99/97 & 96/94 & 100/100 & & 100/98 \\
		bmr4302 & 115 & 107 & 82/80 & 91/91 & 100/99 & & 99/99 \\
		bmr4670 & 120 & 102 & 83/81 & 88/87 & 98/97 & & 98/97 \\
		bmr4353 & 126 & 98 & 83/80 & 90/90 & 99/93 & & 95/95 \\
		bmr4207 & 158 & 148 & 82/81 & 88/85 & 100/97 & & 99/99 \\
		bmr4318 & 215 & 191 & 84/75 & 74/70 & 100/98 & & 98/98 \\
		\hline
	\end{tabular}
	\flushleft
	{\footnotesize ${}^1$ Number of assignable spin systems in the BMRB data.}
\end{table}

It can be seen that LIAN-1 achieves an assignment performance that generally exceeds that of both CISA and IPASS, particularly on recall (with the aforementioned caveat about the IPASS results, which may be overestimated by virtue of not including random noise). In particular, our algorithm performs strongly on \texttt{bmr4353}, which is particularly challenging due to the large number of Proline residues (which, as mentioned in \ref{section:seq walking}, do not yield \fifteenn--\oneh{N} interactions in the chosen spectra). 

One notable exception is the smallest protein, \texttt{bmr4391}. The reason for the lower performance on this particular instance appears to be that LIAN-1 finds an assignment of significantly higher likelihood than the ground-truth assignment (at least according to the generative model we selected). In fact, under the chosen generative model, our relaxation-based algorithm finds the optimal solution to the original (un-relaxed) Problem 1, which is small enough in this instance to solve directly. This is a useful reminder that the ground-truth assignment (often determined manually) may not maximize likelihood under our probabilistic model.

\subsubsection{FLYA simulated framework}

The simulated framework described for the protein SH2 in \cite{ass:flya} was used to generate noisy peak lists, as validation of the peak list graph model described in \ref{ch-lp-graph}. In particular, artificial peak lists were generated for HSQC, HN(CO)CACB, HNCACB, HNCO, HN(CO)CA, HN(CA)CO, and HNCA spectra at the positions specified by the reference chemical shifts as listed in the corresponding BMRB entry \cite{ass:BMRB}. The measured frequencies for each peak were then randomly shifted by adding white Gaussian noise, with standard deviations of 0.4/4 ppm for \thirteenc{} and \fifteenn atoms, and 0.03/4 ppm for \oneh{N} atoms. Deviations that exceeded 0.4 ppm (for \thirteenc{} and \fifteenn atoms) or 0.04 ppm (for \oneh{N} atoms) were discarded, as per the simulation description in \cite{ass:flya}. This is a best-effort approach at replicating the exact simulation framework in that paper.

The node enumerator described in Appendix \ref{app:graph_construction} was used with only the 4 largest maximal cliques for each connected component considered as a node. The results are summarized in Table \ref{tab:synthetic_sh2}, where we can see that LIAN-1 appears to deliver comparable performance to FLYA. Three scores are presented for the LIAN-1 approach, corresponding to the lowest, average, and highest \% correctness in the assignments over a set of 20 simulations.

\begin{table}[h!]\footnotesize
	\centering
	\caption[Performance of LP on SH2 peak list datasets]{Percentage of correct atom assignments for LIAN-1 and FLYA on simulated SH2 peak list datasets. For LIAN-1 we show the lowest, \textbf{average}, and highest correctness scores achieved over 20 simulations.}
	\label{tab:synthetic_sh2}
	\begin{tabular}{cccccc}
		\hline\\
		\textbf{Protein ID} & \textbf{Length} & \textbf{FLYA} & \textbf{LIAN-1} \\\hline
		SH2 & 114 & 97.2\% & 94.5\%, \textbf{95.4\%}, 97.5\% \\
		\hline
	\end{tabular}
\end{table}

\subsection{Experimental data}

To validate the performance of LIAN on experimental data, we make use of the experimental dataset used by the authors of IPASS in \cite{ass:ipass}. This is a challenging dataset, as there are several missing spin systems. We also note that some of the spin systems that were manually assigned are significantly distinct from the prior (this could be a legitimate biological phenomenon, as shielding effects resulting from the specific electronic environment of the protein shift the resonance frequencies of atoms). LIAN-2 (Problem \ref{lp-prob-r2}) was used throughout, with $\lambda=5$. Results are summarized in Table \ref{tab:ipass_lp}, which shows the number of correctly assigned residues alongside the number of assigned residues for each methodology. 

\begin{table}[h!]\footnotesize
	\centering
	\caption[Performance of LP on real datasets.]{Accuracy of assignment on four distinct spin system datasets provided by the authors of IPASS. Results for other algorithms were obtained from \cite{ass:ipass}.}
	\label{tab:ipass_lp}
	\begin{tabular}{lccccccc}
		\hline
		\textbf{Protein} & \textbf{Length} & \textbf{Manual}${}^1$ & \textbf{Spins}${}^2$ & \textbf{MARS} & \textbf{IPASS} & \textbf{C-SDP} & \textbf{LIAN-2} \\ \hline
		TM1112 & 89 & 83 & 81/85 & 55/63 & 71/72 & 50/85 & 74/81 \\
		CASKIN & 67 & 54 & 47/48 & 23/25 & 29/39 & 27/48 & 36/44 \\
		VRAR & 72 & 60 & 47/47 & 6/17 & 30/37 & 19/47 & 29/51\\
		HACS1 & 74 & 61 & 48/61 & 15/16 & 37/50 & 19/61 & 39/53\\\hline
	\end{tabular}
	\flushleft
	{\footnotesize For each method, we show the results in the form \#Correct/\#Assigned, where \#Correct is the number of correctly assigned residues and \#Assigned is the total number of assigned residues.\\
		${}^1$ Number of manually assigned residues in the BMRB file.\\
		${}^2$ Correct/Total available spin systems, where spin systems are considered correct (i.e. not artifacts) if they were manually assigned by NMR practitioners. These numbers are taken from \cite{ass:ipass}. We note that there are meaningful differences between the spin system values and the chemical shifts available publicly on BMRB, so the first number should be interpreted as an upper bound on the number of potentially assignable residues.}
\end{table}

We see that LIAN-2 delivers state-of-the-art performance on this dataset in terms of recall (albeit at the expense of lower precision relative to IPASS). However, we note that the precision-recall threshold can be easily tuned by adjusting the threshold score in the dummy nodes, and that once an assignment is produced, validation of assigned spin systems can be made more easily by referring to the residue-level likelihood scores for debugging (which is why higher recall was selected for in our thresholds). 

An important observation provided by these experiments is that LIAN-2 is particularly useful when datasets are of poor quality. In fact, we observe that the final solution in all these experiments reused several of the spin systems in multiple positions, which would not have been possible under the standard formulation presented in Problem \ref{lp-prob-r0}. This illustrates the importance of correctly characterizing the quality of the dataset through appropriate constraints on the problem.
\section{Conclusion}\label{sec:conclusion}

This paper introduced a novel formulation of the spectral assignment problem in NMR as a constraint satisfaction problem. More specifically, we formulate it as a constrained shortest path problem, for which near-optimal solutions can be found via linear programming relaxations. This approach has significant advantages over existing approaches, as it treats spectral assignment as a global optimization problem, without the need for intermediate steps (such as spin system creation) which can lead to information loss. Furthermore, the approach is amenable to multiple probabilistic characterizations and could therefore accommodate complex characterizations of the generative model for the data (such as aminoacid-specific chemical shift correlations), which would simply result in different edge weights for the assignment graph introduced in Section \ref{methodology}. This approach could also straightforwardly accommodate other interactions often useful for assignment, such as the existence of hydrogen-hydrogen interactions in NOE spectra, through additional linear costs in the objective function.

Testing of our approach with a simplistic generative model on both simulated and experimental data showed state-of-the-art performance. For synthetic, spin system data, our methodology's performance matched or surpassed the best performing algorithms (IPASS, \cite{ass:ipass} and CISA, \cite{Cisa}), with a notable exception where our algorithm found a higher likelihood assignment than the reference assignment, under our probabilistic model. For experimental, spin system data, our methodology improved upon state-of-the-art. For the higher dimensional problem of peak list data, our preliminary studies indicate performance on par with state-of-the-art algorithm, FLYA.

Our reformulation of the assignment problem permits a more realistic basis for assessment of complete automated structure determination, including ambiguous assignment and constraint methods \cite{ARIAweb}.

\begin{acknowledgements}
A.S. was partially supported by NSF BIGDATA award IIS-1837992, NIH/NIGMS award 1R01GM136780-01, award FA9550-17-1-0291 from AFOSR, the
Simons Foundation Math+X Investigator Award, and the Moore Foundation Data-Driven Discovery Investigator Award. DC was supported by NIH GM-117212.
\end{acknowledgements}

%
\section*{Conflict of interest}

The authors declare that they have no conflict of interest.

\section*{Data and code availability}

Data and preliminary (non-production) code used in simulations and tests is available in the author's repository at \url{https://github.com/fsbravo/lipras}.

\bibliographystyle{spmpsci}      
\bibliography{bibliography}   
\newpage
\appendix
\section{ - Grouping Peaks}\label{app:graph_construction}

\begin{figure}[!h]
    \centering
    \includegraphics[width=1\columnwidth]{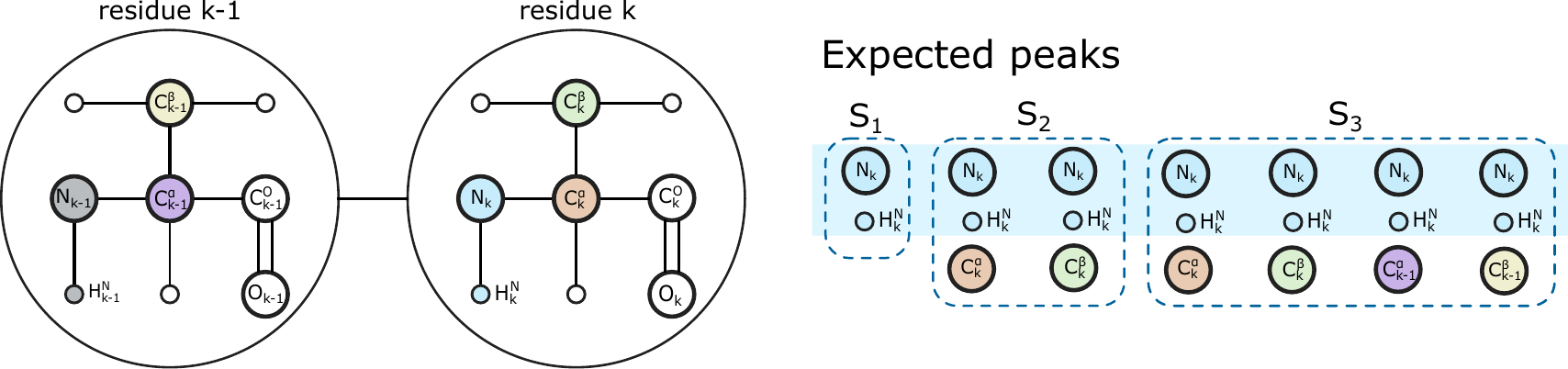}
    \caption{With three NMR experiments (often HSQC, HNCACB, and HN(CO)CACB) we generally expect 7 distinct peaks for each base \fifteenn--\oneh{N} pair in a residue, $k$. These peaks must be consistent - that is, the frequencies assigned to the same atom by two different peaks must be approximately the same up to some experimental tolerance. In principle, there should be $n$ sets of such 7 peaks, one for each residue.}
    \label{fig:free_enumeration}
\end{figure}

As we mentioned in Section \ref{sec:peak_groupings}, grouping consistent peaks together is a crucial step in the graph creation process for $\mathcal{G}=(\mathcal{V},\mathcal{E})$. One would wish the enumeration of valid assignments to be as thorough as possible. We can effectively enumerate peak groupings to construct nodes in $\mathcal{G}$ by matching measured and expected peaks in a self-consistent way. In particular, we expect a specific set of peaks due to \fifteenn--\oneh{N} from residue $k$ (see Figure \ref{fig:free_enumeration} for a standard example with three experiments) where the values of these peaks in $\mathbb{R}^3$ along certain dimensions are consistent. If there are $n$ residues, we should have $n$ sets of such expected peaks. Therefore, each layer in $\mathcal{G}=(\mathcal{V},\mathcal{E})$ in principle should have $n$ nodes, although in practice there are more nodes due ambiguities. 

The notion of consistency can help significantly simplify the enumeration process (which would otherwise result in an exponential number of nodes). In order to efficiently enumerate consistent peak groupings, we do the following. Let $\mcal{S}_1, \ldots, \mcal{S}_{L}$ be collections of \emph{measured} peak lists corresponding to different heteronuclear experiments, i.e. $\cup_{l=1}^L\mcal{S}_l:=[p_1, \ldots, p_{m_2}]$. In the case of Figure~\ref{fig:free_enumeration}, $L=3$, as we have peaks from three experiments. 
Now from these $m_2$ experimental peaks we form all combinations of seven peaks that each consists of one peak from $\mathcal{S}_1$, two peaks from $\mathcal{S}_2$, and four peaks from $\mathcal{S}_3$ using the following criteria.
    \begin{itemize}
    \item For any pair of $p_u, p_v$ in a combination of seven peaks, 
    \begin{align*}
	    \vert p_u(1)-p_v(1)\vert&\leq \delta_1 \\
	    \vert p_u(2)-p_v(2)\vert&\leq \delta_2.
	\end{align*}
    This means that the frequencies of the seven peaks in the \fifteenn--\oneh{N} dimension have to coincide up to tolerance $\delta_1,\delta_2$.
    \item Furthermore, for a combination of seven peaks, let $p_u, p_v$ be the two peaks in $\mathcal{S}_2$. These peaks should coincide with two of the peaks in $\mathcal{S}_3$ (denoted $p_i,p_j$) up to tolerance $\delta_3$, i.e.
    \begin{align*}
    \vert p_u(3)-p_i(3)\vert&\leq \delta_3 \\
    \vert p_v(3)-p_j(3)\vert&\leq \delta_3
    \end{align*}
    along the $\aC$ dimension.
\end{itemize}

\section{ - Atom cost}\label{app:atom-cost}

Recall that we defined the cost of an atom, $a$, under a given set of assigned observations, $\{x_l\}_{l=1}^{o_a}$ as

\begin{defn}[Atom cost]
	The cost associated with atom $a$, with a normally distributed prior $\mcal{N}(\mu_a, \sigma_a)$, and $o_a$ observations $\{x_l^a\}_{l=1}^{o_a}$ defined by the peak grouping, also assumed to be normally distributed around the true frequency, $\mu$, according to $\mcal{N}(\mu, \sigma_l)$ is defined as
	\begin{equation}
	\text{cost}\left(a, \{x_l^a\}_{l=1}^{o_a}\right) \triangleq -\log\mathbb{E}_{\mu\sim \mathcal{N}(\mu_a, \sigma_a)}\left[\prod_{l=1}^{o_a}f(x_l^a\mid \mu, \sigma_l)\right].
	\end{equation}
	where $f(\cdot\mid u, v)$ is the Gaussian density with mean $u$ and standard deviation $v$.
\end{defn}

This is Definition 1 in the main text. Note that the term inside the expectation is a product of $o_a$ univariate Gaussian probability density functions. Furthermore, expanding the expectation, we note that

\begin{align}
    \mathbb{E}_\mu\left[\prod_{l=1}^{o_a}f(x_l^a\mid \mu, \sigma_l)\right] &= \int_{-\infty}^{+\infty}f(\mu\mid \mu_a, \sigma_a)\prod_l^{o_a}f(x_l^a\mid \mu, \sigma_l)d\mu \\
    &=\int_{-\infty}^{+\infty}f(\mu\mid \mu_a, \sigma_a)\prod_l^{o_a}f(\mu\mid x_l^a, \sigma_l)d\mu
\end{align}
by symmetry. Using a standard result regarding the product of univariate Gaussian PDFs (see, e.g., \cite{gaussianproduct}), we can write

\begin{align}
    \mathbb{E}_\mu\left[\prod_{l=1}^{o_a}f(x_l^a\mid \mu, \sigma_l)\right] &=\int_{-\infty}^{+\infty}f(\mu\mid \mu_a, \sigma_a)\prod_l^{o_a}f(\mu\mid x_l^a, \sigma_l)d\mu \\
    &=\int_{-\infty}^{+\infty}Z_af(\mu\mid M_a, \Sigma_a)d\mu \\
    &=Z_a
\end{align}
where
\begin{align}
\Sigma_a &= \left(\frac{1}{\sigma_a^2}+\sum_{l=1}^{o_a} \frac{1}{\sigma_l^2}\right)^{-1/2} \\
M_a &= \left(\frac{\mu_a}{\sigma_a^2}+\sum_{l=1}^{o_a}\frac{x_l}{\sigma_l^2}\right)\Sigma^2_{a} \\
Z_a & =\frac{1}{(2\pi)^{o_a/2}}\sqrt{\frac{\Sigma_a^2}{\sigma_a^2\prod_{l=1}^{o_a}\sigma_l^2}}\exp\left[-\frac{1}{2}\left(\frac{\mu_a^2}{\sigma_a^2}+\sum_{l=1}^{o_a}\frac{x_l^2}{\sigma_l^2}-\frac{M_a^2}{\Sigma_a^2}\right)\right].
\end{align}

We see that this choice of cost function is therefore computationally advantageous, as the desired expectation is a simple function of the observations, $\{x_l\}_{l=1}^{o_a}$ and of the distributional parameters of the prior, $(\mu_a, \sigma_a)$ and experiments, $\{\sigma_l\}_{l=1}^{o_a}$. That said, it is certainly not the only cost function that one could use. As an example, we could instead solve a maximum likelihood problem for each peak grouping that would assign the highest likelihood frequency to each atom, given the prior and the observations. The exploration of alternative cost functions is left for future work.

\section{ - Statistical Typing}\label{app:statistical-typing}

Statistical typing is a process that happens both during the node and edge creation steps. In particular, we want to avoid the creation of nodes and edges which are too unlikely to constitute a valid assignment. The way we action on this notion is to define a threshold below which we would rather have a \emph{null} assignment than the assignment induced by the relevant nodes. This threshold also determines the cost of the edges to (and from) the \emph{dummy} nodes, which are therefore the highest cost edges in the graph.

For all simulations in this paper, we use the following definition:
\begin{defn}[Atom cost threshold]
	The maximum allowable cost associated with atom $a$, with an expected frequency, $\mu$, distributed according to the normally distributed prior $\mcal{N}(\mu_a, \sigma_a)$, and a total of $o_a$ expected observations is given by:
	\begin{equation}
	    \text{threshold}\left(a\right) \triangleq \text{cost}(a, \{w_l^a\}_{l=1}^{o_a})
	\end{equation}
	where
	\begin{equation}
	    w_l^a = \mu_a +\delta\sigma_a + (-1)^{l+1}\delta\sigma_l.
	\end{equation}
\end{defn}

That is, we define the maximum allowable cost for atom $a$ by setting $\{x^a_l\}_{l=1}^{o_a}$ in Definition \ref{def:atom cost} to $\{w^a_l\}_{l=1}^{o_a}$, which constitute an adversarial realization of the observations. In this realization, the mean of the observations is $\approx \delta$ standard deviations away from the prior mean, and the observations are split into two clusters, $2\delta$ experimental standard deviations apart.

\end{document}